\documentclass[10pt,twocolumn,letterpaper]{article}
\usepackage[pagenumbers]{cvpr} 
\usepackage{amsmath,amssymb,booktabs,graphicx,tabularx,array}
\usepackage{algorithm,algpseudocode}
\definecolor{cvprblue}{rgb}{0.21,0.49,0.74}
\usepackage[pagebackref,breaklinks,colorlinks,allcolors=cvprblue]{hyperref}
\hypersetup{pdfauthor={},pdftitle={Unified Residual Translation for Paired Image Restoration},
 pdfsubject={Anonymous submission},pdfkeywords={image translation, residual learning, image restoration}}
\def\confName{CVPR}
\def\confYear{2027}
\newcommand{\E}{\mathbb{E}}
\newcommand{\F}{\mathcal{F}}
\newcommand{\G}{\mathcal{G}}
\newcommand{\clip}{\operatorname{clip}}
\newcommand{\meanabs}[1]{\lVert #1\rVert_{1,\mathrm{mean}}}

\title{Unified Residual Translation for Paired Image Restoration}
\author{\textbf{Joyanta Jyoti Mondal}\textsuperscript{1,4,*},
\textbf{Mridul Banik}\textsuperscript{2,4,*},
\textbf{Ibne Farabi Shihab}\textsuperscript{4},
\textbf{Md. Shifatul Ahsan Apurba}\textsuperscript{3,4}
\\
\textsuperscript{1}University of Delaware,
\textsuperscript{2}Indiana University Indianapolis,\\
\textsuperscript{3}University of Alabama at Birmingham,
\textsuperscript{4}BRAC University\\
\small{
\textsuperscript{\textbf{*}}Equal Contribution.
\textbf{Correspondence:}
\href{mailto:joyanta@udel.edu}{joyanta@udel.edu} and \href{mailto:mapurba@uab.edu}{mapurba@uab.edu}
}
}

\begin{document}
\maketitle

\begin{abstract}
Paired image restoration spans changes in imaging contrast, sensor modality, and spatial resolution. These settings differ in their input channels and target grids, raising the question of how much of the prediction pipeline must be specialized to each task. We present the Unified Residual Translator (URT), which separates fixed channel and scale mapping from learned residual prediction. A deterministic reference uses channel averaging, replication, or bicubic interpolation to match the target dimensions, while a NAFNet encoder-decoder predicts an additive correction from the complete input. Separately trained instances share the prediction rule, $\ell_1$ objective, and optimizer across MRI contrast synthesis, visible-to-infrared translation, SAR-to-optical translation, and real-camera super-resolution. Across six task and scale settings, URT achieves the highest PSNR and SSIM among the evaluated baselines on IXI, SEN1-2, and RealSR $\times3$ and $\times4$. It reaches 30.774\,dB on IXI and 13.478\,dB on SEN1-2, with luminance PSNR of 33.596, 28.742, and 27.342\,dB at the three RealSR scales. On LLVIP, it ranks second in PSNR and SSIM. A study of four spectral alternatives identifies complementary benefits of fitted references and bandwise residual gains. Together, these results support explicit dimensional alignment and learned residual correction as a common formulation for heterogeneous paired imaging tasks.
\end{abstract}

\section{Introduction}
\label{sec:intro}

Paired image restoration recovers a target image from a corresponding observation. Across imaging systems, this relationship can involve a change in contrast, sensor modality, or spatial resolution. MRI contrast synthesis maps neighboring source slices to a target contrast; visible-to-infrared translation changes both channel structure and appearance; SAR-to-optical translation maps a single-channel observation to RGB; and real-camera super-resolution increases spatial resolution. These tasks share paired supervision, but their inputs and outputs occupy different image spaces. A common predictor must accommodate these differences while preserving the information available in the source.

Residual learning expresses image prediction as a reference plus a correction~\cite{vdsr,edsr,nafnet}. Applying this principle across heterogeneous tasks requires an explicit choice of reference. An identity shortcut cannot directly connect tensors with different channel counts or spatial dimensions, and matching dimensions alone does not resolve a change in appearance. This motivates separating two operations within the predictor. A fixed mapping reconciles channel count and spatial scale, while a learned branch models the transformation from the complete source observation. We investigate whether this separation provides an effective common construction across modality translation and super-resolution.

\begin{figure*}[t]
\centering
\includegraphics[width=.92\textwidth]{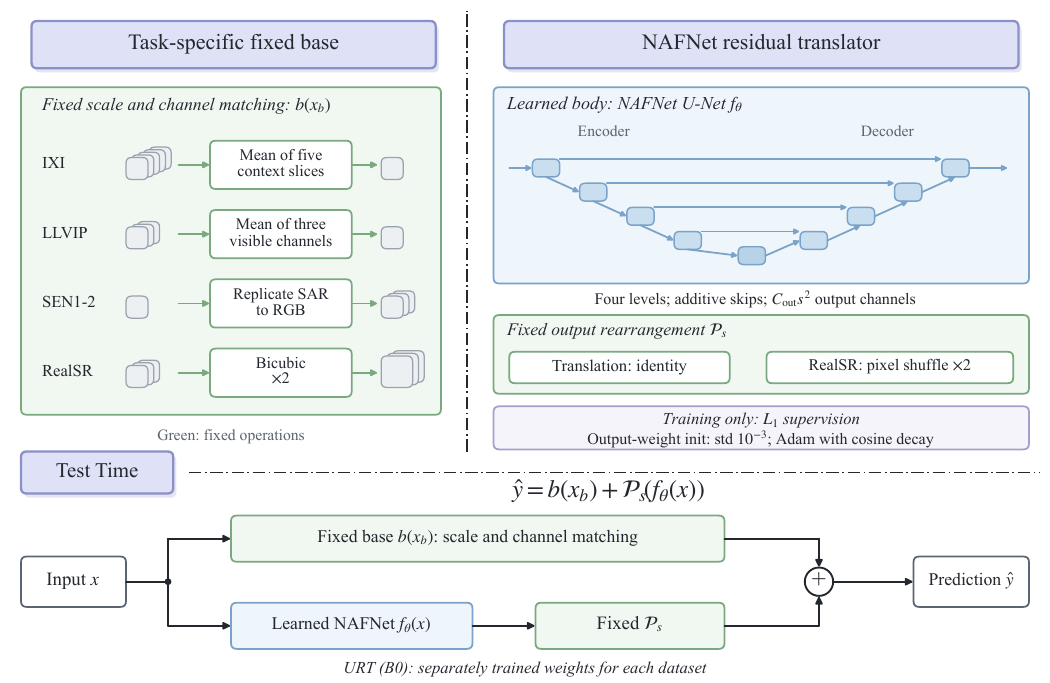}
\caption{\textbf{Unified Residual Translator.} A fixed reference reconciles source and target dimensions, and a NAFNet residual body learns the correction from the complete input. The reference averages the five IXI context slices or the three LLVIP visible channels, replicates SEN1-2 SAR to RGB, or applies bicubic upsampling for RealSR. Pixel shuffle maps the super-resolution residual onto the target grid; the diagram illustrates the $\times2$ case. The diagram's fixed base $b(x_b)$ denotes $B_s(x)$ in Eq.~\ref{eq:b0-base}; adding the two branches gives the raw predictor $F_\theta(x)$ in Eq.~\ref{eq:model}. Green denotes fixed operations and blue the learned body. Training uses paired supervision and separately learned weights for each dataset.}
\label{fig:overview}
\end{figure*}

We introduce the Unified Residual Translator (URT), which combines a deterministic reference with a NAFNet encoder-decoder~\cite{nafnet} that predicts the remaining correction. The reference uses an arithmetic channel mean, channel replication, or bicubic interpolation according to the input and output dimensions. These operations are confined to the reference branch. The learned branch receives every source channel, including all five MRI context slices, and produces a correction on the target grid. For super-resolution, pixel shuffle~\cite{espcn} converts output channels into spatial samples. Small output-head initialization centers the initial prediction on the reference, and a common $\ell_1$ objective supervises reconstruction. Figure~\ref{fig:overview} summarizes the construction. Each dataset has separately trained weights and its own dimensional settings, width, and schedule, while retaining the same prediction rule and body design.

The same formulation provides a setting for studying a more structured reference. We compare URT with four spectral variants that replace the fixed mapping with a Fourier-domain map fitted to training pairs and optionally introduce learned bandwise residual gains. The residual body definition and task schedules are held consistent. This comparison examines where fitted references and residual modulation improve reconstruction, and where a fixed dimensional mapping is sufficient.

We evaluate IXI~\cite{ixi}, LLVIP~\cite{llvip}, SEN1-2~\cite{sen12}, and RealSR~\cite{realsr} across six task and scale settings, with 22 baseline runs. 
The spectral study complements these comparisons by identifying when fitted references and residual gains benefit the same predictor design.

Our contributions are:
\begin{itemize}
\item A common residual formulation that explicitly reconciles source and target dimensions through a fixed reference while learning the transformation from the complete input.
\item An evaluation across four datasets and six task and scale settings, comparing the same predictor design and reconstruction objective with task-specific baselines.
\item An analysis of fitted spectral references and residual gains that identifies their task-dependent benefits relative to the fixed-reference construction.
\end{itemize}

\section{Related work}
\label{sec:related}

\paragraph{Residual restoration.}
VDSR~\cite{vdsr} and EDSR~\cite{edsr} learn residual corrections for super-resolution, while RCAN~\cite{rcan} and SwinIR~\cite{swinir} introduce attention-based restoration architectures. NAFNet~\cite{nafnet} and Restormer~\cite{restormer} provide encoder-decoder designs for image restoration. URT uses a NAFNet body within an explicit fixed-reference construction. Channel mapping and output rearrangement allow the same residual rule to operate when source and target differ in channel count or spatial resolution, while the learned branch retains the complete observation.

\paragraph{Spectral and model-based prediction.}
GFNet~\cite{gfnet}, adaptive Fourier neural operators~\cite{afno}, and fast Fourier convolution~\cite{ffc} incorporate frequency-domain processing into learned representations. FFTformer~\cite{fftformer} and Fourmer~\cite{fourmer} apply spectral modeling to restoration. Deep Wiener Deconvolution~\cite{dwdn} and USRNet~\cite{usrnet} combine model-based estimates with learned reconstruction. Our spectral comparisons examine a related choice within a common residual framework: a frozen channel map fitted in radial Fourier bands using ridge regression~\cite{ridge}, optionally followed by learned bandwise scaling of the residual.

\paragraph{Image translation.}
pix2pix~\cite{pix2pix} learns conditional adversarial translation, while CycleGAN~\cite{cyclegan} introduces cycle-consistency constraints. Pyramid pix2pix (BCI)~\cite{bci} adds Gaussian-pyramid reconstruction supervision. Palette~\cite{palette}, Brownian bridge models~\cite{bbdm}, and image-to-image Schr\"odinger bridges~\cite{i2sb} formulate translation through generative paths. Recent work makes the role of residuals and path construction explicit. DRDD~\cite{drdd} separates stochastic noise diffusion from deterministic residual translation, and RDBM~\cite{rdbm} uses residuals to modulate diffusion-bridge perturbations. Dual-approx Bridge~\cite{dualbridge} learns forward and reverse approximators, while ResFlow~\cite{resflow} models degradation through deterministic flow paths. FlowSR~\cite{flowsr} combines rectified flow with consistency learning for single-step super-resolution, and conditional consistency models~\cite{ccm} extend consistency prediction to translation and enhancement. MeanFlow~\cite{meanflow} learns interval-average velocities, and Conditional MeanFlow (CMF)~\cite{racmf} conditions this prediction on a source image. URT directly learns an additive correction with an $\ell_1$ objective, evaluated through complementary reconstruction and perceptual metrics~\cite{blau,freirich}.

\paragraph{Imaging tasks.}
MRI synthesis includes ResViT~\cite{resvit} and SynDiff~\cite{syndiff}. LLVIP~\cite{llvip} supports visible-to-infrared translation studied by ThermalGen~\cite{thermalgen} and conditional consistency models~\cite{ccm}. SEN1-2~\cite{sen12} provides paired SAR and optical imagery, with S2O-TDN~\cite{s2otdn} addressing their translation. RealSR~\cite{realsr} and DRealSR~\cite{dreal} study real-camera super-resolution. OR-Net~\cite{ornet} decomposes features into frequency components and aggregates them through spatially adaptive operations; MPF-Net~\cite{mpfnet} provides another task-specific comparison. We organize evaluation by dataset and scale to account for differences in pairing, preprocessing, and scoring.

\section{Method}
\label{sec:method}

URT combines a fixed reference in target coordinates with a learned correction conditioned on the complete input.

\subsection{A fixed reference in target coordinates}
\label{sec:b0-reference}

Let $x\in\mathbb{R}^{C_{\mathrm{in}}\times H\times W}$ and $y\in\mathbb{R}^{C_{\mathrm{out}}\times sH\times sW}$ denote a preprocessed, spatially aligned pair. We use $s=1$ for modality translation and $s\in\{2,3,4\}$ for super-resolution. Define $\mathcal U_1$ as the identity and $\mathcal U_s$ for $s>1$ as bicubic interpolation to the target grid with pixel-center alignment. A fixed channel map $A\in\mathbb{R}^{C_{\mathrm{out}}\times C_{\mathrm{in}}}$ acts at each spatial location:
\begin{equation}
A_{oc}=
\begin{cases}
\mathbf{1}[o=c], & C_{\mathrm{in}}=C_{\mathrm{out}},\\[1mm]
1/C_{\mathrm{in}}, & C_{\mathrm{in}}\ne C_{\mathrm{out}}.
\end{cases}
\label{eq:b0-channel-map}
\end{equation}
The reference image is
\begin{equation}
B_s(x)=A\,\mathcal U_s(x).
\label{eq:b0-base}
\end{equation}
For unequal channel counts, the map averages the input channels and repeats that mean across the required output channels. Table~\ref{tab:b0-mappings} gives the evaluated mappings. These operations introduce no fitted parameters.

\begin{table}[t]
\small
\caption{\textbf{Task instantiations of the fixed reference.} Channel means are arithmetic. The learned branch always receives the full input.}
\label{tab:b0-mappings}
\setlength{\tabcolsep}{4pt}
\begin{tabularx}{\columnwidth}{@{}lcc>{\raggedright\arraybackslash}X@{}}
\toprule
Task & $C_{\mathrm{in}}\!\to\!C_{\mathrm{out}}$ & $s$ & Reference $B_s(x)$\\
\midrule
IXI & $5\to1$ & 1 & Five-slice mean\\
LLVIP & $3\to1$ & 1 & Visible-channel mean\\
SEN1-2 & $1\to3$ & 1 & SAR replicated to RGB\\
RealSR & $3\to3$ & $2,3,4$ & Bicubic RGB upsampling\\
\bottomrule
\end{tabularx}
\end{table}

Channel reduction affects only the reference: the learned branch receives every input channel and can predict distinct corrections in each target channel.

\subsection{Residual prediction from the full input}
\label{sec:b0-residual}

The body $f_\theta$ maps $x$ to $C_{\mathrm{out}}s^2$ channels on the input grid. Let $\mathcal P_s$ denote pixel shuffle~\cite{espcn}, which rearranges those channels onto the target grid, with $\mathcal P_1$ the identity. The raw prediction is
\begin{equation}
F_\theta(x)=B_s(x)+\mathcal P_s\!\left(f_\theta(x)\right).
\label{eq:model}
\end{equation}
Both branches meet in target coordinates. For modality translation, the residual learns changes in contrast and channel appearance. For super-resolution, feature computation remains on the low-resolution grid until the output rearrangement. The correction can modify intensity, color, and spatial detail throughout the reference.

We instantiate $f_\theta$ with a NAFNet encoder-decoder~\cite{nafnet}. A $3\times3$ stem produces $w$ channels. Four encoder levels use widths $(w,2w,4w,8w)$ and block counts $(1,1,1,28)$, followed by one bottleneck block at width $16w$. Each level downsamples with a stride-2 $2\times2$ convolution. Decoder levels upsample using a $1\times1$ convolution and pixel shuffle, add the corresponding encoder features, and apply one block. The blocks use channel normalization, multiplicative SimpleGate, and simplified channel attention. A final $3\times3$ convolution produces $C_{\mathrm{out}}s^2$ channels. Bottom and right reflection padding make input dimensions divisible by 16; the output is cropped to $H\times W$ before $\mathcal P_s$. We use $w=48$ for IXI and $w=32$ otherwise. Here $f_\theta$ denotes the learned correction branch; Eq.~\ref{eq:model} defines its image-level addition to the reference.

\paragraph{Initialization.}
Output-head weights are independent $\mathcal N(0,\sigma^2)$ samples with $\sigma=10^{-3}$, and the bias is zero. NAFNet's internal residual scales also start at zero. For a fixed vectorized $3\times3$ decoder-feature patch $h_p$ and an output-head filter $W_c$, the initial correction $o_{c,p}=W_c^\top h_p$ satisfies
\begin{equation}
\E[o_{c,p}\mid h_p]=0,\qquad
\E[o_{c,p}^2\mid h_p]=\sigma^2\|h_p\|_2^2.
\label{eq:init-moments}
\end{equation}
These moments follow directly from the independent, zero-mean weights. Initialization therefore centers the prediction on the fixed reference, with correction variance determined by the output-head scale and feature energy.

\subsection{Residual supervision}
\label{sec:b0-training}

For a minibatch $\mathcal B$ of jointly augmented pairs, training minimizes mean absolute reconstruction error:
\begin{equation}
\mathcal L_{\mathcal B}(\theta)=
\frac{1}{|\mathcal B|}\sum_{(x,y)\in\mathcal B}
\left\|F_\theta(x)-y\right\|_{1,\mathrm{mean}},
\label{eq:loss}
\end{equation}
where $\|z\|_{1,\mathrm{mean}}=\|z\|_1/\operatorname{numel}(z)$. The corresponding target in body-output coordinates is
\begin{equation}
r_s^\star(x,y)=\mathcal P_s^{-1}\!\left(y-B_s(x)\right).
\label{eq:b0-target}
\end{equation}
Because pixel shuffle permutes the same scalar entries,
\begin{equation}
\left\|F_\theta(x)-y\right\|_{1,\mathrm{mean}}
=
\left\|f_\theta(x)-r_s^\star(x,y)\right\|_{1,\mathrm{mean}}.
\label{eq:b0-residual-loss}
\end{equation}
Equation~\ref{eq:b0-residual-loss} expresses the shared learning problem: the reference defines the correction, and the rearrangement defines its coordinates. The implementation computes Eq.~\ref{eq:loss} directly on the unclipped image prediction.

Each dataset has separately trained parameters. Optimization uses Adam~\cite{adam} with moments $(0.9,0.999)$ and cosine learning-rate decay to $10^{-7}$. Paired augmentation preserves spatial correspondence. The term \emph{unified} denotes the prediction rule, residual body design, and reconstruction objective; dimensions, widths, schedules, and learned weights follow each task configuration. Algorithm~\ref{alg:b0} summarizes training and checkpoint selection.

\begin{algorithm}[t]
\caption{URT training and checkpoint selection}
\label{alg:b0}
\small
\begin{algorithmic}[1]
\Require Training pairs; task dimensions, width $w$, update budget $T$, learning rate $\eta_0$; validation data for translation
\Ensure Selected parameters $\theta^\star$
\State Define $A$, $B_s$, and $\mathcal P_s$ using Eqs.~\ref{eq:b0-channel-map}--\ref{eq:model}
\State Initialize the NAFNet residual body
\State Set block residual scales and head bias to zero
\State Draw head weights from $\mathcal N(0,10^{-6})$
\State Initialize Adam and $q_{\mathrm{best}}\gets-\infty$
\For{$t=0,\ldots,T-1$}
  \State Obtain the next jointly augmented minibatch $\mathcal B$
  \State Compute $F_\theta(x)$ using Eq.~\ref{eq:model}
  \State Compute $\mathcal L_{\mathcal B}$ using Eq.~\ref{eq:loss}
  \State Update $\theta$ with Adam at cosine-decayed rate $\eta_t$
  \If{$s=1$ and a training epoch has ended}
    \State $q\gets$ mean validation PSNR using Eq.~\ref{eq:b0-infer} with $\theta^\star=\theta$
    \If{$q>q_{\mathrm{best}}$}
      \State $\theta^\star\gets\theta$; $q_{\mathrm{best}}\gets q$
    \EndIf
  \EndIf
\EndFor
\If{$s>1$}
  \State $\theta^\star\gets\theta$ at the fixed final update
\EndIf
\State \Return $F_{\theta^\star}$ with the task inference procedure
\end{algorithmic}
\end{algorithm}

\subsection{Inference}
\label{sec:b0-inference}

Translation uses the checkpoint with the highest mean validation PSNR. With horizontal reflection $\mathcal H$, evaluation averages the original and reflected predictions before clipping:
\begin{equation}
\hat y=
\clip_{[0,1]}\!\left(
\frac{F_{\theta^\star}(x)+
\mathcal H F_{\theta^\star}(\mathcal H x)}{2}
\right).
\label{eq:b0-infer}
\end{equation}
The raw predictor in Eq.~\ref{eq:model} uses one body evaluation; Eq.~\ref{eq:b0-infer} uses two.

RealSR uses the final checkpoint of its fixed schedule. Inference evaluates $256$-pixel low-resolution tiles with $24$-pixel overlap, averages predictions uniformly over overlapping target-grid supports, and clips the assembled image to $[0,1]$. Each tile is evaluated once, without flip averaging.

\section{Experiments}
\label{sec:experiments}

\begin{table*}[!t]
\centering
\small
\setlength{\tabcolsep}{2.5pt}
\caption{\textbf{Comparison on image translation and real-camera super-resolution.} All metrics are test-set scores. PSNR is in dB; RealSR PSNR/SSIM use luminance. Parameters are in millions. \textbf{Bold} and \underline{underlining} mark the best and second-best values within each dataset and scale. $\uparrow$: higher is better; $\downarrow$: lower is better. $\dagger$ marks a paper-based reimplementation. See Section~\ref{sec:baselines} for implementation details.}
\label{tab:comparison}
\begin{minipage}[t]{.485\textwidth}
\centering
\textbf{(a) Modality translation}\par\smallskip
\begin{tabular*}{\linewidth}{@{\extracolsep{\fill}}lrrrr@{}}
\toprule
Model & PSNR $\uparrow$ & SSIM $\uparrow$ & LPIPS $\downarrow$ & Params. $\downarrow$\\
\midrule
\multicolumn{5}{l}{\emph{IXI}}\\
URT (ours) & \textbf{30.774} & \textbf{0.957} & \underline{0.032} & \textbf{38.292}\\
pix2pix~\cite{pix2pix} & \underline{28.747} & \underline{0.934} & \textbf{0.030} & 54.412\\
SynDiff~\cite{syndiff} & 22.154 & 0.700 & 0.117 & \underline{39.725}\\
\midrule
\multicolumn{5}{l}{\emph{LLVIP}}\\
URT (ours) & \underline{13.126} & \underline{0.359} & 0.569 & 17.111\\
pix2pix~\cite{pix2pix} & 12.002 & 0.283 & \underline{0.513} & 54.410\\
CycleGAN~\cite{cyclegan} & 11.840 & 0.283 & 0.585 & \underline{11.372}\\
BCI~\cite{bci} & \textbf{13.270} & \textbf{0.376} & \textbf{0.482} & 11.377\\
Consistency~\cite{ccm} & 12.666 & 0.255 & 0.698 & 128.916\\
CMF~\cite{racmf} & 13.093 & 0.241 & 0.634 & \textbf{6.149}\\
\midrule
\multicolumn{5}{l}{\emph{SEN1-2}}\\
URT (ours) & \textbf{13.478} & \textbf{0.229} & \underline{0.844} & 17.111\\
S2O-TDN~\cite{s2otdn}$^\dagger$ & \underline{12.831} & \underline{0.208} & \textbf{0.625} & \underline{9.166}\\
CMF~\cite{racmf} & 10.575 & 0.044 & 0.945 & \textbf{6.150}\\
\bottomrule
\end{tabular*}
\end{minipage}\hfill
\begin{minipage}[t]{.485\textwidth}
\centering
\textbf{(b) Real-camera super-resolution}\par\smallskip
\begin{tabular*}{\linewidth}{@{\extracolsep{\fill}}lrrrr@{}}
\toprule
Model & PSNR $\uparrow$ & SSIM $\uparrow$ & LPIPS $\downarrow$ & Params. $\downarrow$\\
\midrule
\multicolumn{5}{l}{\emph{RealSR $\times2$}}\\
URT (ours) & 33.596 & 0.920 & 0.146 & 17.115\\
EDSR~\cite{edsr} & \underline{33.821} & \underline{0.922} & \underline{0.145} & \textbf{1.370}\\
RCAN~\cite{rcan} & 33.467 & 0.917 & 0.150 & 15.445\\
OR-Net~\cite{ornet}$^\dagger$ & \textbf{33.975} & \textbf{0.924} & \textbf{0.141} & 9.135\\
MPF-Net~\cite{mpfnet}$^\dagger$ & 33.760 & 0.920 & 0.147 & \underline{5.959}\\
CMF~\cite{racmf} & 30.766 & 0.856 & 0.162 & 6.152\\
\midrule
\multicolumn{5}{l}{\emph{RealSR $\times3$}}\\
URT (ours) & \textbf{28.742} & \textbf{0.812} & \underline{0.382} & 17.119\\
EDSR~\cite{edsr} & 28.254 & 0.782 & 0.441 & \textbf{1.554}\\
RCAN~\cite{rcan} & \underline{28.684} & 0.790 & \textbf{0.379} & 15.629\\
OR-Net~\cite{ornet}$^\dagger$ & 28.494 & \underline{0.801} & 0.396 & 10.057\\
MPF-Net~\cite{mpfnet}$^\dagger$ & 26.457 & 0.731 & 0.563 & \underline{6.076}\\
\midrule
\multicolumn{5}{l}{\emph{RealSR $\times4$}}\\
URT (ours) & \textbf{27.342} & \textbf{0.767} & \underline{0.473} & 17.125\\
EDSR~\cite{edsr} & 26.590 & 0.726 & 0.535 & \textbf{1.518}\\
RCAN~\cite{rcan} & \underline{27.178} & 0.738 & \textbf{0.461} & 15.592\\
OR-Net~\cite{ornet}$^\dagger$ & 27.089 & \underline{0.756} & 0.491 & 11.348\\
MPF-Net~\cite{mpfnet}$^\dagger$ & 24.891 & 0.641 & 0.664 & \underline{6.240}\\
\bottomrule
\end{tabular*}
\end{minipage}
\end{table*}

\subsection{Datasets and evaluation protocol}
\label{sec:protocol}

All experiments use images from the publicly released IXI~\cite{ixi}, LLVIP~\cite{llvip}, SEN1-2~\cite{sen12}, and RealSR~\cite{realsr} benchmarks for computational training and evaluation, under the partitions and preprocessing below.

\paragraph{IXI.}
T1-to-T2 synthesis on IXI~\cite{ixi} uses 25/5/10 training/validation/test subjects, with 100 slices per subject at $256\times256$ resolution. Inputs stack five neighboring T1 slices. T2 is affinely registered to T1 using mutual information. Each volume is divided by its own mean, and scaling maxima are fitted on training subjects. The input window selects brain slices below the vertex. For evaluation, prediction and reference are separately mean-normalized and then divided by the reference maximum.

\paragraph{LLVIP.}
Visible-to-infrared translation uses the official 3,463-image test split~\cite{llvip}. Full frames are resized to $256\times256$ without preserving aspect ratio. Validation uses 305 subsampled images from training scenes 07, 11, and 15; all remaining frames from those scenes are excluded from training.

\paragraph{SEN1-2.}
SAR-to-RGB translation uses 209/23/26 training/validation/test season/ROI scene tokens~\cite{sen12}. Up to 40 evenly spaced patches per scene yield 8,360/920/1,040 pairs at native $256\times256$ resolution. SAR is decoded as one eight-bit channel and the target is RGB.

\paragraph{RealSR.}
We evaluate RealSR V3~\cite{realsr} at scales $2$, $3$, and $4$. The $\times2$ configuration uses 406 training pairs and the official 100-image test set, with paired $48\times48$ low-resolution crops and $96\times96$ targets for 200,000 updates. Its final checkpoint is evaluated without a validation split. The $\times3$ and $\times4$ results extend the comparison to larger target grids and are reported separately by scale.

\paragraph{Implementation.}
Translation uses batch size 8, with initial learning rates $10^{-4}$ for IXI and $2\times10^{-4}$ for LLVIP and SEN1-2. The RealSR $\times2$ configuration uses batch size 32 and initial rate $10^{-4}$. IXI uses flips and affine augmentation with probability 0.7; LLVIP uses flips; SEN1-2 uses flips and right-angle rotations; and RealSR uses flips and rotations on paired crops.

\paragraph{Metrics.}
PSNR and SSIM~\cite{ssim} measure reconstruction, and LPIPS~\cite{lpips} measures perceptual distance. Higher PSNR/SSIM and lower LPIPS are preferred. LLVIP PSNR/SSIM uses floating-point predictions; LPIPS uses quantized infrared replicated into three channels. SEN1-2 predictions are quantized before scoring. Translation SSIM uses the default uniform window. RealSR quantizes predictions to eight bits, converts to luminance, and removes a border equal to the scale for PSNR-Y/SSIM-Y. Its SSIM uses a Gaussian window with standard deviation 1.5 and population covariance. RealSR LPIPS uses full RGB images. LPIPS uses AlexNet features. Parameter counts describe model size, with fewer parameters indicating a smaller model.

\subsection{Retrained comparisons}
\label{sec:baselines}

\begin{table*}[!t]
\centering
\small
\setlength{\tabcolsep}{4pt}
\caption{\textbf{Residual-family analysis on four task settings.} B0 is URT; B1 uses a fitted spectral reference; B2 adds unit-initialized gains; B2r and B3 use random and cross-fitted gain profiles with profile regularization. Body designs and task schedules are shared. \textbf{Bold} and \underline{underlining} mark the best and second-best distinct values in each column, including ties. $\uparrow$: higher is better; $\downarrow$: lower is better. Panels (a) and (b) report test metrics. Panel (c) combines translation validation PSNR and RealSR $\times2$ test PSNR-Y for mixed rank; test rank uses test PSNR for all four settings. Each rank orders the five configurations; mean ranks weight the four settings equally. PSNR is in dB.}
\label{tab:variants}
\textbf{(a) Reconstruction on the test sets}\par\smallskip
\begin{tabular*}{\textwidth}{@{\extracolsep{\fill}}lrrrrrrrr@{}}
\toprule
 & \multicolumn{2}{c}{IXI} & \multicolumn{2}{c}{LLVIP} & \multicolumn{2}{c}{SEN1-2} & \multicolumn{2}{c}{RealSR $\times2$}\\
\cmidrule(lr){2-3}\cmidrule(lr){4-5}\cmidrule(lr){6-7}\cmidrule(l){8-9}
Model & PSNR $\uparrow$ & SSIM $\uparrow$ & PSNR $\uparrow$ & SSIM $\uparrow$ & PSNR $\uparrow$ & SSIM $\uparrow$ & PSNR-Y $\uparrow$ & SSIM-Y $\uparrow$\\
\midrule
URT (B0) & \textbf{30.774} & \textbf{0.957} & 13.126 & 0.359 & 13.478 & 0.229 & \textbf{33.596} & \textbf{0.920}\\
B1 & \underline{30.677} & \underline{0.940} & 12.958 & 0.353 & \textbf{13.667} & 0.228 & 33.467 & 0.914\\
B2 & 30.663 & 0.939 & \textbf{13.355} & 0.366 & 13.492 & \textbf{0.236} & \underline{33.489} & \underline{0.915}\\
B2r & 30.629 & 0.938 & \underline{13.332} & \underline{0.377} & 13.407 & \underline{0.235} & 29.451 & 0.758\\
B3 & 30.389 & 0.937 & 13.317 & \textbf{0.382} & \underline{13.520} & \textbf{0.236} & 33.260 & 0.909\\
\bottomrule
\end{tabular*}\par
\medskip
\textbf{(b) LPIPS on the test sets}\par\smallskip
\begin{tabular*}{.72\textwidth}{@{\extracolsep{\fill}}lrrr@{}}
\toprule
Model & LLVIP $\downarrow$ & SEN1-2 $\downarrow$ & RealSR $\times2$ $\downarrow$\\
\midrule
URT (B0) & \textbf{0.5690} & \underline{0.8443} & 0.1458\\
B1 & \underline{0.5800} & \textbf{0.8069} & \textbf{0.1416}\\
B2 & 0.5859 & 0.9687 & \underline{0.1450}\\
B2r & 0.5858 & 0.9531 & 0.3553\\
B3 & 0.5945 & 0.9282 & 0.1477\\
\bottomrule
\end{tabular*}\par
\medskip
\textbf{(c) Validation scores and cross-task summaries}\par\smallskip
\begin{tabular*}{\textwidth}{@{\extracolsep{\fill}}lrrrrrr@{}}
\toprule
 & \multicolumn{3}{c}{Validation PSNR $\uparrow$} & Test PSNR-Y $\uparrow$ & \multicolumn{2}{c}{Mean rank $\downarrow$}\\
\cmidrule(lr){2-4}\cmidrule(lr){5-5}\cmidrule(l){6-7}
Model & IXI & LLVIP & SEN1-2 & RealSR $\times2$ & Mixed & Test\\
\midrule
URT (B0) & 29.487 & \textbf{18.832} & \textbf{14.387} & \textbf{33.596} & \textbf{1.75} & \underline{2.50}\\
B1 & 29.575 & 18.188 & 14.170 & 33.467 & 3.75 & 2.75\\
B2 & \textbf{29.589} & 18.147 & 14.247 & \underline{33.489} & \underline{3.00} & \textbf{2.25}\\
B2r & \underline{29.587} & 18.237 & \underline{14.327} & 29.451 & \underline{3.00} & 4.00\\
B3 & 29.186 & \underline{18.294} & 14.272 & 33.260 & 3.50 & 3.50\\
\bottomrule
\end{tabular*}
\end{table*}

The comparison includes 22 baseline runs across three modality-translation tasks and three RealSR scales. Models are evaluated with task-specific preprocessing and scorers, while conditioning, objectives, optimizer conventions, and training endpoints follow their respective configurations. URT uses the checkpoint and inference procedures in Section~\ref{sec:b0-inference}. Results are compared within each dataset and scale.

Official implementations are used for pix2pix~\cite{pix2pix}, CycleGAN~\cite{cyclegan}, Pyramid pix2pix (BCI)~\cite{bci}, SynDiff~\cite{syndiff}, and the consistency model~\cite{ccm}. EDSR~\cite{edsr} and RCAN~\cite{rcan} use the BasicSR architectures with the $\ell_1$/Adam training setup. pix2pix and CycleGAN retain their default losses and optimizer settings, while SynDiff receives the central T1 slice. S2O-TDN~\cite{s2otdn}, OR-Net~\cite{ornet}, and MPF-Net~\cite{mpfnet} are paper-based reimplementations, marked $\dagger$.

\begin{figure*}[!t]
\centering
{\small\textbf{(a) IXI T1-to-T2}\par}
\includegraphics[width=.833\textwidth,trim=0 180bp 0 0,clip]{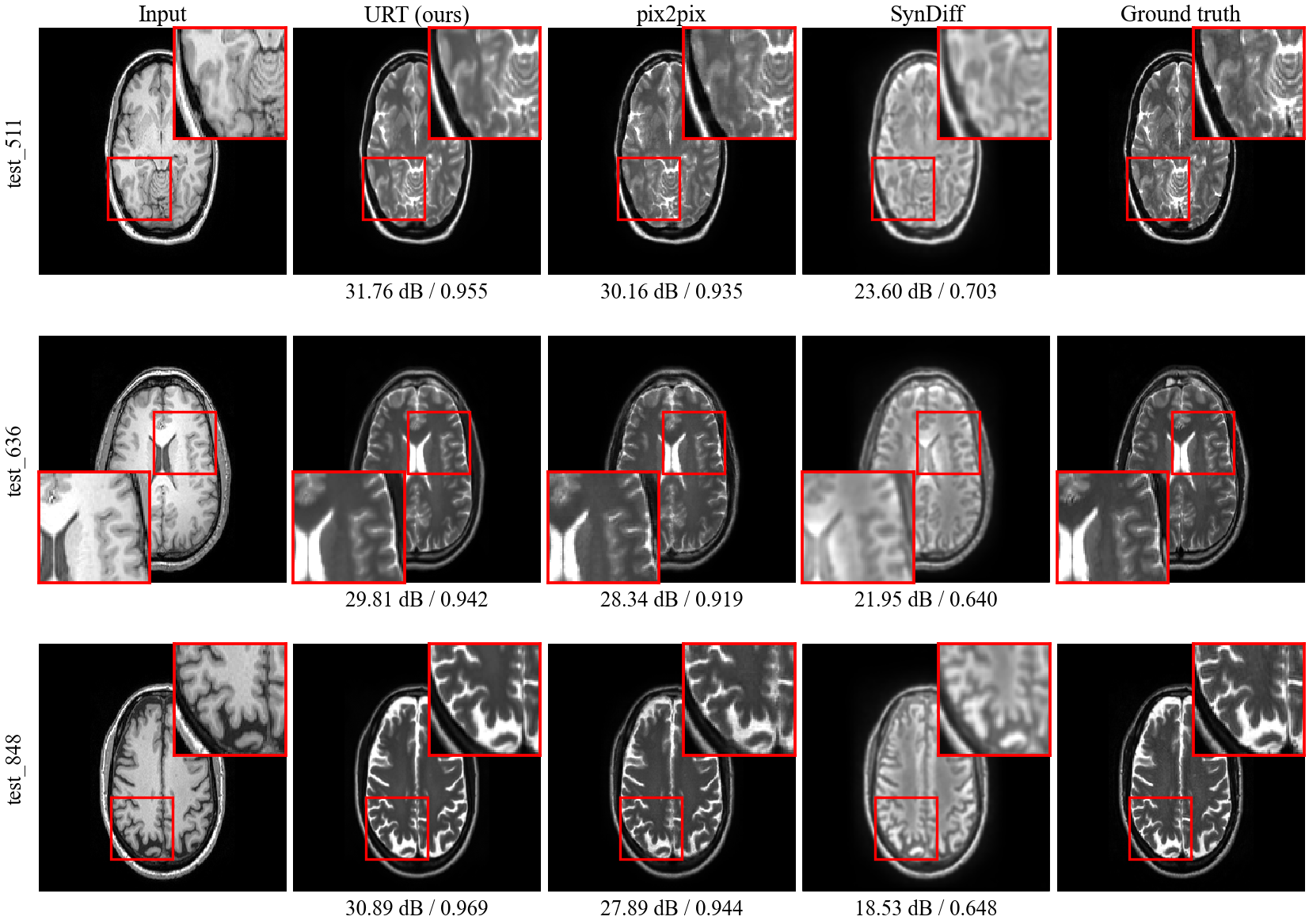}\par
\smallskip
{\small\textbf{(b) LLVIP visible-to-infrared}\par}
\includegraphics[width=\textwidth,trim=0 155bp 0 0,clip]{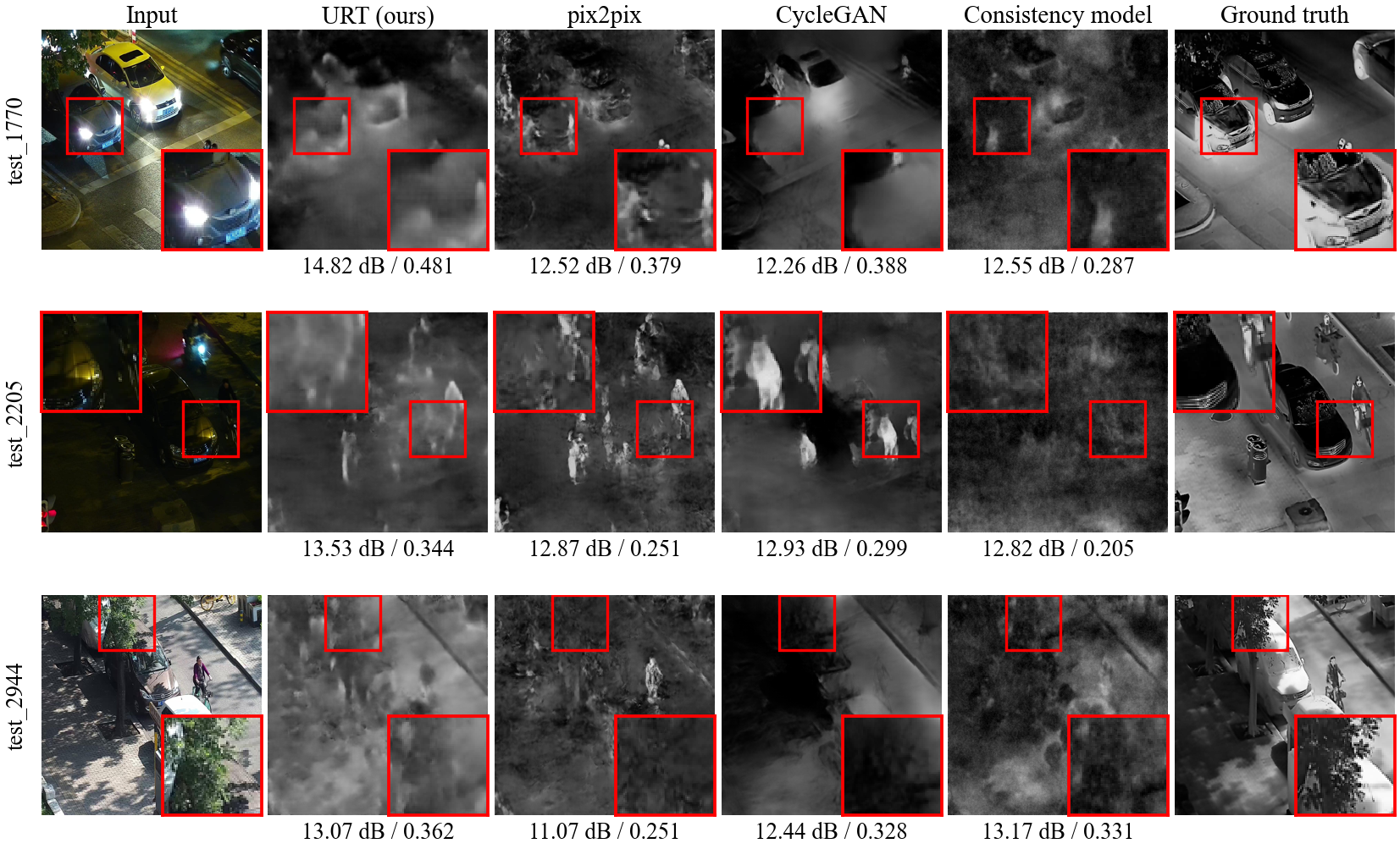}\par
\smallskip
{\small\textbf{(c) SEN1-2 SAR-to-optical}\par}
\includegraphics[width=.667\textwidth,trim=0 220bp 0 0,clip]{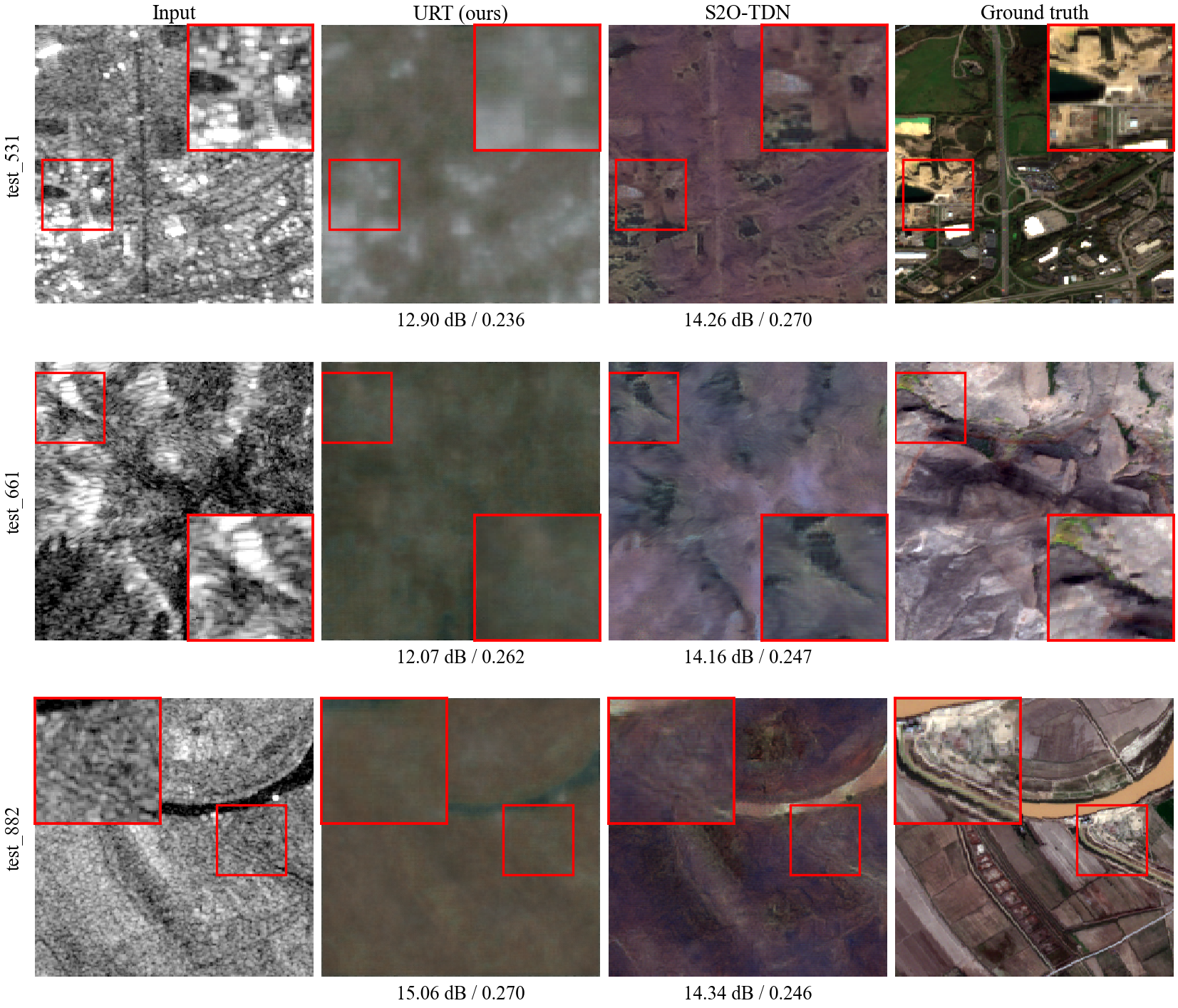}\par
\smallskip
{\small\textbf{(d) RealSR $\times2$}\par}
\includegraphics[width=\textwidth,trim=0 118bp 0 0,clip]{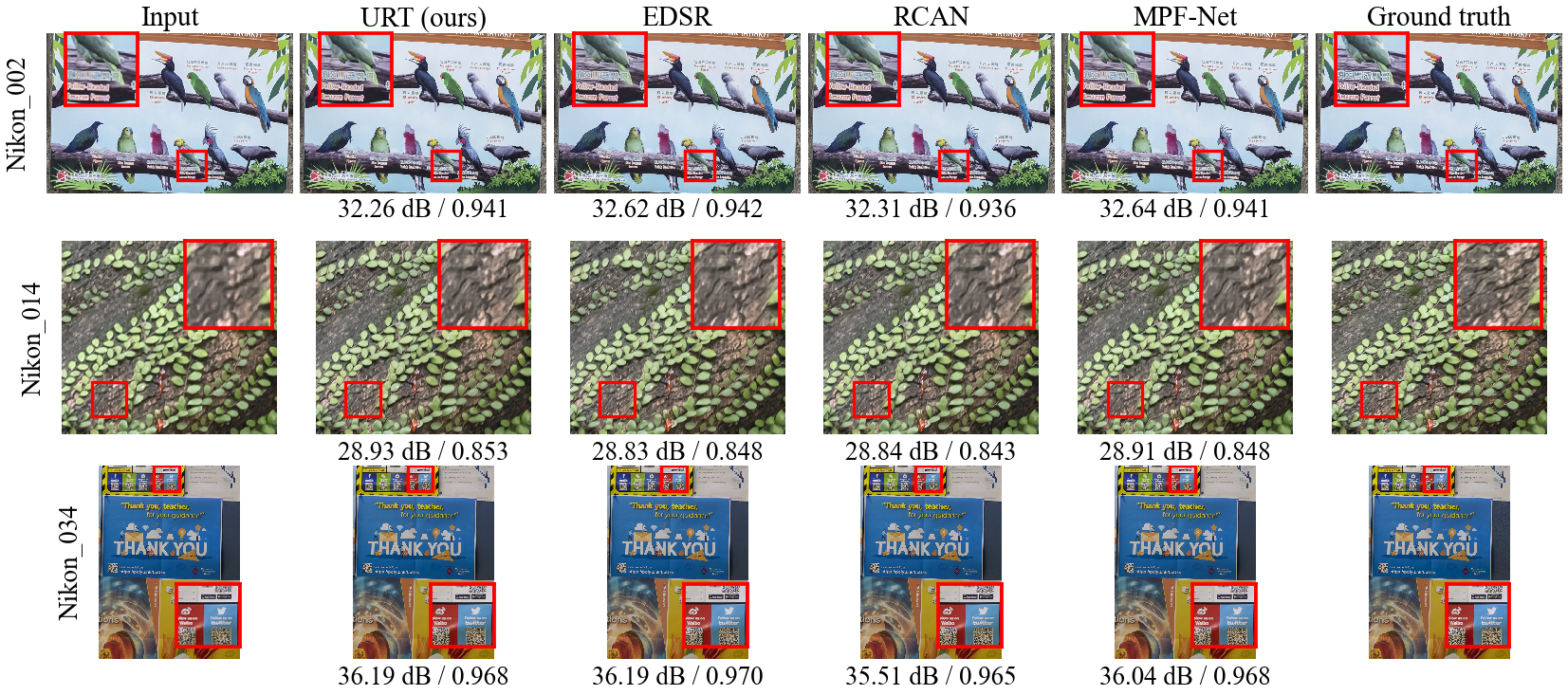}
\caption{\textbf{Representative qualitative comparisons across four tasks.} The displayed examples are IXI slice 511, LLVIP frame 1770, SEN1-2 patch 531, and RealSR Nikon\_002, taken in the fixed order of the preselected random samples. Red boxes identify a target-selected detail window enlarged consistently across methods. Values are per-example PSNR (dB) and SSIM; RealSR uses luminance metrics. IXI output columns share the target-based display scale, with separate scaling for the central T1 input. LLVIP uses floating predictions on $[0,1]$; SEN1-2 and RealSR use the quantized outputs of their scoring protocols. The displayed RealSR input is enlarged by nearest-neighbor replication.}
\label{fig:qualitative}
\end{figure*}

\subsection{Quantitative results}
\label{sec:results}

Table~\ref{tab:comparison} reports reconstruction, perceptual distance, and model size. URT gives the highest PSNR and SSIM on IXI, SEN1-2, and RealSR $\times3$/$\times4$, and the second-highest values on LLVIP. The taskwise results distinguish these reconstruction gains from the ordering under LPIPS.

\paragraph{MRI contrast synthesis.}
URT achieves 30.774\,dB PSNR and 0.957 SSIM on IXI, improving on pix2pix by 2.027\,dB and 0.023 SSIM. Its LPIPS is 0.032, compared with 0.030 for pix2pix, and its 38.292\,M parameters make it the smallest evaluated IXI model. SynDiff reaches 22.154\,dB with the central T1 slice as input. URT also leads the residual family in IXI PSNR and SSIM (Table~\ref{tab:variants}), supporting the fixed context-slice reference for contrast synthesis.

\paragraph{Visible-to-infrared translation.}
URT reaches 13.126\,dB PSNR and 0.359 SSIM on LLVIP, ranking second in both metrics and lying within 0.144\,dB of BCI. BCI leads this comparison with 13.270\,dB, 0.376 SSIM, and 0.482 LPIPS. Relative to CMF, URT improves PSNR by 0.033\,dB and SSIM by 0.118, while reducing LPIPS from 0.634 to 0.569. CMF has the smallest parameter count, 6.149\,M, separating model size from the reconstruction ordering.

\paragraph{SAR-to-optical translation.}
On the 26-scene-token SEN1-2 test set, URT achieves 13.478\,dB PSNR and 0.229 SSIM, exceeding the S2O-TDN reimplementation by 0.647\,dB and 0.021 SSIM. Relative to CMF, the gains are 2.903\,dB and 0.185 SSIM, with lower LPIPS (0.844 versus 0.945). S2O-TDN achieves the lowest LPIPS, 0.625, illustrating the distinct ordering of reconstruction and perceptual distance.

\paragraph{Real-camera super-resolution.}
At $\times2$, URT achieves 33.596\,dB PSNR-Y and 0.920 SSIM-Y, exceeding RCAN by 0.129\,dB and CMF by 2.830\,dB. OR-Net leads this scale with 33.975\,dB, 0.924 SSIM-Y, and 0.141 LPIPS, followed by EDSR. At $\times3$ and $\times4$, URT has the highest PSNR-Y and SSIM-Y, reaching 28.742/0.812 and 27.342/0.767, respectively. Its PSNR-Y gains over the next-best RCAN are 0.058 and 0.164\,dB. RCAN has the lowest LPIPS at these two scales, with URT second. The same residual formulation therefore supports all three enlargement factors while preserving distinct reconstruction and perceptual rankings.

\subsection{Qualitative comparisons}
\label{sec:qualitative}

Figure~\ref{fig:qualitative} compares representative methods on the three translation tasks and RealSR $\times2$. Each row shows the first item from a fixed random test sample, with the same target-selected detail window applied to every displayed method. Per-example scores accompany the images; the complete dataset-level comparison, including the additional baselines and scales, appears in Table~\ref{tab:comparison}.

\section{Analysis}
\label{sec:analysis}

\subsection{Fixed and fitted references}
\label{sec:variants}

We examine the reference and residual parameterization on IXI, LLVIP, SEN1-2, and RealSR $\times2$, retaining the body design and task schedules. B0 denotes URT. B1 replaces $B_s$ in Eq.~\ref{eq:model} with an affine spectral reference. Let $z=[\mathcal U_s(x);\mathbf1]$ append a constant channel to the input on the target grid, and let $\F$ denote the unnormalized two-dimensional Fourier transform. The fitted reference satisfies
\begin{equation}
\F(B_{\mathrm{spec}}(x))(k)=W_{b(k)}\,\F(z)(k),
\label{eq:spectral-reference}
\end{equation}
where $b(k)$ assigns frequency $k$ to one of 16 radial bands. Each real matrix $W_b$ is fitted on training input--target coefficients by ridge regression~\cite{ridge}, with unit weights over stored real-FFT bins. The ridge strength is $\lambda_b=10^{-6}(\operatorname{tr}(A_b)/d+10^{-12})$, with $d=C_{\mathrm{in}}+1$. Here $A_b$ sums $\operatorname{Re}(Z_i(k)Z_i(k)^*)$ over training pairs and stored bins in band $b$, where $Z_i(k)=\F(z_i)(k)$ and $*$ denotes conjugate transpose. The fitted reference remains frozen during residual training. B2, B2r, and B3 additionally apply per-output-channel, per-band gains to the target-grid residual $r=\mathcal P_s(f_\theta(x))$:
\begin{equation}
\F(\G_g r)_c(k)=g_{c,b(k)}\,\F(r)_c(k).
\label{eq:app-gate}
\end{equation}
The gated variants predict $B_{\mathrm{spec}}(x)+\G_g r$.

B2 initializes gains to one and applies no gain penalty. B2r initializes a random profile in $[0.05,1]$. B3 uses two-fold cross-fitting on training data: a reference fitted on one fold predicts the other, and residual and target Fourier energies are pooled across both held-out folds within each channel and band. Folds alternate sorted training subject/scene identifiers, or image indices for RealSR. The residual-to-target energy ratios are clipped to $[0,1]$, square-rooted, normalized to unit maximum within each output channel, and floored at 0.05 to form $g_0$; denominators are bounded below by $10^{-12}$. The reference is then refitted on all training pairs and frozen. B2r and B3 add the same penalty $10^{-3}\|g-g_0\|_F^2/(16C_{\mathrm{out}})$ around their respective initial profiles. Gains remain unconstrained during training. B3 versus B2r therefore compares measured and random profiles under the same penalty; comparisons with B2 also change regularization.

Table~\ref{tab:variants} reports the residual-family comparison. URT gives the highest IXI and RealSR $\times2$ reconstruction scores. On SEN1-2, the fitted reference in B1 raises PSNR from 13.478 to 13.667\,dB and lowers LPIPS from 0.8443 to 0.8069. On LLVIP, unit-initialized B2 reaches the highest PSNR, 13.355\,dB, while calibrated B3 reaches the highest SSIM, 0.382; URT retains the lowest family LPIPS, 0.5690. B1 also has the lowest RealSR $\times2$ LPIPS, 0.1416. These outcomes distinguish the roles of the reference and residual modulation across reconstruction and perceptual criteria.

\subsection{Cross-task behavior and computation}
\label{sec:cross-task}

Within the five-configuration residual family, URT ranks first on LLVIP and SEN1-2 validation PSNR and on RealSR $\times2$ test PSNR-Y, yielding a mixed mean rank of 1.75. Using test PSNR for all four settings, B2 has mean rank 2.25, followed by URT at 2.50 and B1 at 2.75. These summaries weight each setting equally and apply to the four-task spectral study.

Table~\ref{tab:comparison} reports model sizes, including the scale-dependent RealSR output heads. Training on one Tesla P100 takes 6.4\,h for IXI, 5.4\,h for LLVIP, 6.7\,h for SEN1-2, and approximately 20\,h for RealSR $\times2$. A B3 family profile averages approximately 41\,ms per $256\times256$ input in fp32 at batch size one over 50 repetitions. This timing includes B3's fitted base and gain operations alongside the residual body shared with URT.

\paragraph{Discussion.}
The results support separating dimensional alignment from learned image synthesis. URT ranks among the top two in both PSNR and SSIM in five of the six dataset--scale settings in Table~\ref{tab:comparison}, using a common residual formulation. Its fixed reference establishes the target dimensions while the learned branch retains all source channels. Equation~\ref{eq:b0-residual-loss} shows that output rearrangement preserves the reconstruction objective exactly, linking modality translation and super-resolution through the same residual-learning problem. The spectral study in Table~\ref{tab:variants} makes reference construction and residual modulation explicit design choices: fixed references are effective on IXI and RealSR, while fitted structure offers complementary benefits for sensor translation. The distinct reconstruction and perceptual rankings motivate complementary objectives. Future work includes joint training across datasets and evaluation with additional seeds and modalities.

\section{Conclusion}
\label{sec:conclusion}

We presented URT, a common residual formulation for paired image translation and real-camera super-resolution. A deterministic channel-and-scale mapping supplies the reference, while a NAFNet body learns the target correction from the complete input. Separately trained instances share the prediction rule and reconstruction objective across four datasets. URT leads the evaluated baseline comparisons in PSNR and SSIM on IXI, SEN1-2, and RealSR $\times3$/$\times4$, ranks second on LLVIP, and achieves 33.596\,dB PSNR-Y on RealSR $\times2$. The spectral analysis identifies complementary benefits of fitted references and residual gains, clarifying the role of reference construction across heterogeneous paired imaging tasks.

\section*{LLM Usage Statement}
Generative AI tools assisted with literature discovery, mathematical exposition, code preparation, numerical analysis, and language editing. ChatGPT and Claude assisted with manuscript organization, LaTeX revision, and checks of mathematical exposition, citations, and numerical consistency. The authors are responsible for the scientific claims, experimental interpretation, and final submission.

{\small
\bibliographystyle{ieeenat_fullname}
\bibliography{main}
}
\newpage
\providecommand{\E}{\mathbb{E}}
\providecommand{\F}{\mathcal{F}}
\providecommand{\G}{\mathcal{G}}
\providecommand{\clip}{\operatorname{clip}}
\providecommand{\tr}{\operatorname{tr}}
\providecommand{\meanabs}[1]{\lVert #1\rVert_{1,\mathrm{mean}}}
\makeatletter
\@ifundefined{proposition}{\newtheorem{proposition}{Proposition}}{}
\@ifundefined{lemma}{\newtheorem{lemma}{Lemma}}{}
\makeatother

\clearpage
\setcounter{figure}{0}
\setcounter{table}{0}
\setcounter{equation}{0}
\renewcommand{\thefigure}{S\arabic{figure}}
\renewcommand{\thetable}{S\arabic{table}}
\renewcommand{\theequation}{S\arabic{equation}}
\renewcommand{\theHfigure}{S\arabic{figure}}
\renewcommand{\theHtable}{S\arabic{table}}
\renewcommand{\theHequation}{S\arabic{equation}}
\begin{center}\large Appendix\end{center}
\appendix

\section{Overview and notation}
\label{app:overview}
The appendix is organized as follows.
Section~\ref{app:related} extends the related-work discussion.
Section~\ref{app:derivations} gives the derivations behind the fixed and fitted references, the residual gains and the NAFNet body.
Section~\ref{app:experimental} documents datasets, scorers, the URT and baseline configurations, hardware and software, efficiency and model selection.
Section~\ref{app:results} reports additional results, and Section~\ref{app:scope} states the scope of the evidence.
Table~\ref{tab:notation} fixes the notation, which follows the main paper, and Table~\ref{tab:experiments} lists the experiments.

\begin{table}[!h]
\centering\small
\caption{Notation. Symbols follow the main paper.}
\label{tab:notation}
\setlength{\tabcolsep}{3pt}
\begin{tabularx}{\columnwidth}{@{}l>{\raggedright\arraybackslash}X@{}}
\toprule
Symbol & Meaning\\
\midrule
$x,\ y$ & Preprocessed, aligned source ($C_{\mathrm{in}}\times H\times W$) and target ($C_{\mathrm{out}}\times sH\times sW$)\\
$s$ & Spatial scale: $1$ for translation, $2,3,4$ for RealSR\\
$\mathcal U_s$ & Identity for $s=1$; bicubic interpolation to the target grid for $s>1$\\
$A$ & Fixed channel map (identity, or channel mean repeated over outputs)\\
$B_s(x)$ & Fixed reference $A\,\mathcal U_s(x)$\\
$f_\theta$ & NAFNet residual body with $C_{\mathrm{out}}s^2$ output channels\\
$\mathcal P_s$ & Pixel shuffle onto the target grid; identity for $s=1$\\
$F_\theta(x)$ & Raw predictor $B_s(x)+\mathcal P_s(f_\theta(x))$\\
$w$ & Body width ($48$ for IXI, $32$ otherwise)\\
$\F$ & Unnormalized two-dimensional DFT, applied per channel\\
$\Omega_+$ & Frequencies stored by the real FFT of a target-grid image\\
$b(k)$ & Radial band of frequency $k$, one of $16$ bands\\
$z$ & Augmented input $[\mathcal U_s(x);\mathbf 1]$, with $d=C_{\mathrm{in}}+1$ channels\\
$W_b$ & Real $C_{\mathrm{out}}\times d$ channel map of band $b$ in the fitted reference\\
$B_{\mathrm{spec}}(x)$ & Fitted spectral reference (B1, B2, B2r, B3)\\
$r$ & Target-grid residual $\mathcal P_s(f_\theta(x))$\\
$\G_g$ & Bandwise residual gain operator with gains $g_{c,b}$ and initial profile $g_0$\\
\bottomrule
\end{tabularx}
\end{table}

\begin{table*}[t]
\centering\small
\caption{Experiments reported in the main paper and this appendix. Every reported model is a single training run.
The residual-family study covers RealSR at $\times2$ only.}
\label{tab:experiments}
\setlength{\tabcolsep}{4pt}
\begin{tabularx}{\textwidth}{@{}>{\raggedright\arraybackslash}p{.17\textwidth}>{\raggedright\arraybackslash}X>{\raggedright\arraybackslash}X>{\raggedright\arraybackslash}X@{}}
\toprule
Experiment & Question & Held fixed & Reported outcome\\
\midrule
Main comparison (main Table~\ref{tab:comparison}) & How does URT compare with task-specific baselines? & Per setting: data split, preprocessing and scorer, across six settings (IXI, LLVIP, SEN1-2, RealSR $\times2,\times3,\times4$) and 22 baseline runs. Baseline objectives, optimizers and training endpoints follow their own configurations. & URT has the highest PSNR and SSIM on IXI, SEN1-2 and RealSR $\times3$ and $\times4$, the second highest on LLVIP, and the fourth highest PSNR-Y among six models at RealSR $\times2$.\\
Residual-family study (main Table~\ref{tab:variants}, Section~\ref{app:family}) & When do a fitted spectral reference and bandwise residual gains help? & Body design, data, task schedule, objective and scorer, on IXI, LLVIP, SEN1-2 and RealSR $\times2$. & URT leads on IXI and RealSR $\times2$; the fitted reference helps on SEN1-2 and residual gains on LLVIP.\\
Single-setting contrasts (Section~\ref{app:sensitivity}) & Which single change to the reference or the gains moves the test score? & Data split, body, width, schedule, seed and scorer, on the four residual-family settings. & The sign depends on the task, and no single change helps on all four settings.\\
RealSR in RGB (Section~\ref{app:rgb}) & Does the luminance ordering hold in full color? & The fixed endpoints of the main comparison. & URT has the highest RGB PSNR at $\times3$ and $\times4$ and in the residual family at $\times2$.\\
Paired RealSR comparison (Section~\ref{app:paired}) & How consistent is the URT versus RCAN gap across test images and cameras? & The $\times3$ and $\times4$ endpoints and their common 100 test images. & The mean gap is larger on Nikon than on Canon images at both scales.\\
PSNR versus LPIPS (Section~\ref{app:worked}) & Do the two metrics rank URT and RCAN the same way on single images? & Full-image scores of the main-comparison endpoints. & They disagree on 32 to 53 of 100 images per scale; Nikon\_002 shows both directions.\\
Qualitative examples (Section~\ref{app:urt-qual}) & What do outputs look like on fixed random test items? & A seeded draw of three test items per setting and a target-guided inset. & Illustrative only; no aggregate number comes from these panels.\\
\bottomrule
\end{tabularx}
\end{table*}

The spectral study compares five configurations on four settings with one seed each.
It shows where a fitted reference or residual gains changed the reported scores, but it does not separate these effects from run-to-run variation.

\section{Extended related work}
\label{app:related}
The main paper's related-work section groups prior work by task.
This section compares the closest methods along four properties that decide how a method relates to URT: what its residual or path represents, what pairing and alignment it needs, where pretrained priors enter, and whether one set of weights serves several tasks.
Table~\ref{tab:rw-restoration} covers residual-style restoration networks and spectral or model-based methods, and Table~\ref{tab:rw-translation} covers translation, residual diffusion and flow, and one-step generative models.

\begin{table*}[t]
\centering\footnotesize
\caption{Residual-style restoration and spectral or model-based methods. ``Weights across tasks'' asks whether one trained model serves several tasks, scales or degradation settings. ``Not stated'' means the cited publication does not report the property; it does not mean the property is absent. LR and HR denote low and high resolution.}
\label{tab:rw-restoration}
\setlength{\tabcolsep}{3pt}
\begin{tabularx}{\textwidth}{@{}>{\raggedright\arraybackslash}p{.12\textwidth}>{\raggedright\arraybackslash}X>{\raggedright\arraybackslash}X>{\raggedright\arraybackslash}p{.14\textwidth}>{\raggedright\arraybackslash}X@{}}
\toprule
Method & Residual or learned operator & Pairing and alignment & Pretrained prior & Weights across tasks\\
\midrule
VDSR~\cite{vdsr} & Network output is added to the interpolated input & Paired LR and HR images; input interpolated to the target size & Not stated & One model shared across scales\\
EDSR~\cite{edsr} & Residual blocks on LR features, upsampling at the end; the paper notes that interpolated input costs time and memory & Paired LR and HR images & The $\times2$ model initializes the $\times4$ model & Multi-scale variant shares the body and keeps scale-specific modules; the single-scale models are separate\\
NAFNet~\cite{nafnet} & Residual blocks inside the network & Paired; evaluated on denoising and deblurring benchmarks & Not stated & Evaluated per benchmark; no multi-task model claimed\\
Fourmer~\cite{fourmer} & Residual image added to the degraded input; Fourier-domain spatial interaction in the network & Paired; deraining, enhancement, dehazing and guided super-resolution & Not stated & Not stated\\
FFTformer~\cite{fftformer} & Frequency-domain attention and feed-forward blocks & Paired; deblurring & Not stated & Deblurring only\\
DWDN~\cite{dwdn} & Wiener deconvolution applied to learned features & Paired; the blur kernel is given (non-blind) & Not stated & Not stated\\
USRNet~\cite{usrnet} & Unfolded half-quadratic splitting with learned modules & LR images synthesized from HR with a known kernel and noise level; both are inputs & Not stated & One model for several scale factors, kernels and noise levels\\
GFNet, AFNO, FFC~\cite{gfnet,afno,ffc} & Learned frequency-domain filters or token mixing inside the network & Recognition and segmentation benchmarks; no image pairs & Not applicable & Not applicable\\
\midrule
URT (ours) & Learned correction added to a fixed reference that matches channels and scale & Paired, spatially aligned & None & Shared design, separate weights per dataset\\
URT, B1 to B3 (ours) & Fixed-form frequency map fitted by ridge regression per radial band, frozen; learned residual; B2 to B3 add bandwise gains & Paired, aligned & None & Separate weights per dataset\\
\bottomrule
\end{tabularx}
\end{table*}

\begin{table*}[t]
\centering\footnotesize
\caption{Translation, residual diffusion and flow, and one-step generative methods. Column conventions follow Table~\ref{tab:rw-restoration}. CMF is shown as described by the cited source; the CMF rows of the main comparison use a local implementation of this design (Table~\ref{tab:baselines}).}
\label{tab:rw-translation}
\setlength{\tabcolsep}{3pt}
\begin{tabularx}{\textwidth}{@{}>{\raggedright\arraybackslash}p{.12\textwidth}>{\raggedright\arraybackslash}X>{\raggedright\arraybackslash}X>{\raggedright\arraybackslash}p{.14\textwidth}>{\raggedright\arraybackslash}X@{}}
\toprule
Method & Residual or path & Pairing and alignment & Pretrained prior & Weights across tasks\\
\midrule
pix2pix~\cite{pix2pix} & Direct conditional generation & Paired & Not stated & Same architecture and objective; trained separately on each dataset\\
BCI~\cite{bci} & Direct generation with Gaussian-pyramid supervision & Paired, registered image pairs & Not stated & Not stated\\
ResViT~\cite{resvit} & Residual connections inside transformer-convolution blocks, not an image residual & Paired multi-contrast MRI & ImageNet-pretrained transformer weights & Unified implementation for varying source and target configurations\\
SynDiff~\cite{syndiff} & Adversarial diffusion with cycle-consistent modules; no residual & Unpaired training; registered volumes for evaluation & None; modules trained jointly without pretraining & Not stated\\
ThermalGen~\cite{thermalgen} & Flow from noise to a thermal latent, conditioned on an RGB latent & Paired RGB and thermal data from several datasets & Pretrained KL-VAE encoder; optional pretrained SiT weights & Joint training across datasets with per-dataset style embeddings\\
S2O-TDN~\cite{s2otdn} & Third-order finite-difference residual structure inside the network & Paired SAR and optical (SEN1-2) & Not stated & Not stated\\
Palette~\cite{palette} & Conditional diffusion on the target; no residual & Four tasks with inputs derived from clean images & Not stated & Generalist multi-task model compared with task-specific models\\
BBDM~\cite{bbdm} & Brownian bridge between the two domains in a latent space & Paired & Pretrained VQGAN & Not stated\\
I2SB~\cite{i2sb} & Bridge between clean and degraded images & Clean and degraded pairs; degradations applied to ImageNet images & Initialized from an unconditional ADM ImageNet checkpoint & Not stated\\
DRDD~\cite{drdd} & Residual $I_{\mathrm{in}}-I_0$ diffused deterministically in a stage separate from the noise stage & Paired for the residual stage; unpaired target images for the noise stage & Optional initialization from natural-image weights & One model for several tasks (all-in-one setting) and single-task settings\\
RDBM~\cite{rdbm} & Residual between the two distributions modulates bridge noise & Paired & Not stated & One model trained on a mixture of five restoration tasks\\
Dual-approx Bridge~\cite{dualbridge} & Brownian bridge with forward and reverse approximators & Paired & Not stated & Not stated\\
ResFlow~\cite{resflow} & Deterministic flow path between degraded and clean images & Paired; no super-resolution task & Not stated & Same architecture for all tasks; trained on each dataset\\
FlowSR~\cite{flowsr} & Rectified flow from LR to HR, one step after consistency learning & LR synthesized with a RealESRGAN-style degradation pipeline; real-world test sets & Pretrained Stable Diffusion 2.1-base, fine-tuned with LoRA & Not stated\\
CCM~\cite{ccm} & No residual; source concatenated with the noised target; single-step sampling & Paired & Not stated & Same method with a per-dataset schedule\\
MeanFlow~\cite{meanflow} & Average velocity along a flow; one step & Not applicable (image generation) & None; trained from scratch & Not applicable\\
CMF~\cite{racmf} & Image-conditioned average velocity; one or few steps & Paired CT images & Not stated & Not stated\\
\midrule
URT (ours) & Learned correction added to a fixed reference that matches channels and scale; one deterministic pass & Paired, spatially aligned & None & Shared design, separate weights per dataset\\
\bottomrule
\end{tabularx}
\end{table*}

\paragraph{What the residual represents.}
Three meanings of ``residual'' appear in the related work, and URT uses the first.
The first is an image-space difference between the target and a reference that is computed from the source.
VDSR~\cite{vdsr} adds its output to the interpolated input, and Fourmer~\cite{fourmer} adds a residual image to the degraded input, which has the same channel count.
URT follows this form but fixes the reference separately from the body: the map $A$ reconciles the channel count, $\mathcal U_s$ the grid, and pixel shuffle places the body output on the target grid, so the same rule covers $C_{\mathrm{in}}\ne C_{\mathrm{out}}$ and $s>1$.
EDSR~\cite{edsr} avoids interpolated input for cost reasons and upsamples features at the end; URT likewise keeps feature computation on the low-resolution grid and adds the reference only in target coordinates.
The second meaning is architectural: ResViT~\cite{resvit}, S2O-TDN~\cite{s2otdn}, EDSR and NAFNet~\cite{nafnet} use residual connections inside the network, which do not define a reference image.
The third is a quantity along a sampling path.
DRDD~\cite{drdd} defines the residual as the input minus the target and diffuses it deterministically in a stage that is separate from its noise stage, and RDBM~\cite{rdbm} uses residuals to modulate the noise of a diffusion bridge.
These methods produce the output by simulating a path, with several network evaluations in the cited experiments (for example 10 steps in RDBM), and most of them start from a noise sample.
URT evaluates one network pass without sampling noise (two passes with flip averaging for translation), and its residual is the plain difference of main Eq.~\eqref{eq:b0-target}, with no path variable.

\paragraph{Pairing and alignment.}
Apart from CycleGAN~\cite{cyclegan} and SynDiff~\cite{syndiff}, which train on unpaired images, the methods above that map a source image to a target image train on paired data, and so does URT.
The pairs differ in origin.
FlowSR~\cite{flowsr} and I2SB~\cite{i2sb} synthesize degraded inputs from clean images, and USRNet~\cite{usrnet} does the same with a known kernel.
URT is trained on the captured RealSR pairs without a synthetic degradation model.
The methods also differ in what alignment they assume.
BCI~\cite{bci} supervises a Gaussian pyramid of registered pairs, and SynDiff trains without pairs and evaluates on registered volumes.
DRDD~\cite{drdd} is the closest case of mixed supervision, since its noise stage uses only unpaired target images.
URT assumes spatially aligned pairs, as stated in Section~\ref{app:scope}: its residual is defined per pixel on the target grid, so misaligned or unpaired inputs fall outside the claims.

\paragraph{Pretrained priors.}
Several methods inherit a prior from large-scale pretraining: ResViT starts its transformer from ImageNet weights, I2SB~\cite{i2sb} from an unconditional ImageNet diffusion checkpoint, BBDM~\cite{bbdm} works in the latent space of a pretrained VQGAN, ThermalGen~\cite{thermalgen} conditions on a pretrained KL-VAE encoder, and FlowSR~\cite{flowsr} fine-tunes a pretrained Stable Diffusion model.
DRDD offers such an initialization as an option.
A pretrained prior can change both what a method synthesizes and how much training data it needs; our comparison isolates neither effect.
URT uses no pretrained prior in the predictor.
The retrained comparators start from random initialization, except the teacher network in the contrastive term of MPF-Net~\cite{mpfnet} (Table~\ref{tab:baselines}).
LPIPS~\cite{lpips} uses pretrained AlexNet weights, but only as a metric.

\paragraph{Weight sharing across tasks.}
Several of the closest methods use one set of weights for several tasks, scales or configurations: VDSR and the multi-scale variant of EDSR across scales, USRNet across scale factors, kernels and noise levels, ResViT across source and target configurations, ThermalGen across datasets through style embeddings, DRDD and RDBM across restoration tasks, and the generalist model of Palette~\cite{palette}.
ResFlow~\cite{resflow} is a closer analogue to URT in this respect: it applies one architecture to all of its tasks and trains it on each dataset.
URT shares its design across tasks, namely the prediction rule, the body, the objective and the optimizer, but it trains separate weights for every dataset and scale.
It therefore makes no claim about a single model that serves several tasks; joint training across datasets is left to future work.

\paragraph{Spectral and model-based methods.}
GFNet~\cite{gfnet}, AFNO~\cite{afno} and FFC~\cite{ffc} learn frequency-domain filters or token mixing by gradient descent inside recognition and segmentation networks, and FFTformer~\cite{fftformer} and Fourmer apply frequency-domain operations inside restoration networks.
DWDN~\cite{dwdn} and USRNet~\cite{usrnet} use an explicit degradation model: DWDN performs Wiener deconvolution on learned features given the blur kernel, and USRNet takes the kernel and noise level as inputs.
The fitted reference of B1 to B3 differs from all of these.
It is a linear map between input and target channels, estimated once from training pairs by regularized least squares in 16 radial bands (Section~\ref{app:base}), and then frozen; it needs no blur kernel, and the body learns the remaining correction.
It resembles a Wiener-type estimator, although we do not analyze this connection.
The spectral study of the main paper compares this fitted reference with the fixed reference of URT; it does not compare against the learned spectral networks above.

\paragraph{One-step conditional generators.}
CCM~\cite{ccm}, FlowSR and CMF~\cite{racmf} produce an image in one evaluation from a noise sample and a conditioning image, with MeanFlow~\cite{meanflow} as the underlying one-step formulation.
URT also uses one evaluation but has no sampling noise, and it adds its output to a reference where these models predict a velocity or a clean image.
The CMF rows of the main comparison use a local implementation of the one-step conditional MeanFlow design, retrained for each dataset, with a fixed noise seed at inference (Section~\ref{app:retrained}); the comparison with CMF therefore tests this implementation and not the cited method as published.

\section{Derivations and implementation}
\label{app:derivations}

This section derives the properties of the references, gains and body that the main paper uses.
Section~\ref{app:fixed-fourier} shows that URT's fixed reference is a special case of the spectral reference.
Section~\ref{app:base} states the objective that the fitted reference of B1 to B3 minimizes, and Section~\ref{app:gradients} gives the residual gains, their gradients and the meaning of the calibration statistic.
Section~\ref{app:body} specifies the NAFNet body and the distribution of its initial output.
The lemmas and propositions are exact statements about the implemented models; the one heuristic is marked as such.

\subsection{The fixed reference in the Fourier domain}
\label{app:fixed-fourier}
The main paper defines URT's reference spatially, $B_s(x)=A\,\mathcal U_s(x)$, while the implementation applies channel maps to real-FFT coefficients.
The two agree because a frequency-independent channel map commutes with the transform.

\begin{lemma}[Channel maps commute with the DFT]
\label{lem:commute}
Let $u\in\mathbb R^{C\times M\times N}$ and let $G\in\mathbb R^{C'\times C}$ act at every spatial location.
Then $\F(Gu)(k)=G\,\F(u)(k)$ for every frequency $k$.
\end{lemma}
\paragraph{Proof.}
For each output channel $o$, $(Gu)_o=\sum_c G_{oc}u_c$.
The DFT is linear and acts on each channel separately, so $\F((Gu)_o)=\sum_c G_{oc}\F(u_c)$.
\hfill$\square$

Applying the lemma to $z=[\mathcal U_s(x);\mathbf1]$ with $G=[A,\mathbf 0]$ gives
\begin{equation}
\F(B_s(x))(k)=[A,\mathbf 0]\,\F(z)(k)\qquad\text{for all }k.
\label{eq:b0-fourier}
\end{equation}
URT's reference is therefore the spectral reference of the main paper's Eq.~\eqref{eq:spectral-reference} with $W_b=[A,\mathbf 0]$ in every band.
The zero final column removes the constant channel, whose transform is nonzero only at $k=0$.
Because $A$ is real, Eq.~\eqref{eq:b0-fourier} preserves conjugate symmetry, so computing it on the stored coefficients $\Omega_+$ and inverting gives the same real image as the spatial formula.
The fitted reference differs from URT only in that $W_b$ is estimated from data and may differ across bands.

\subsection{Fitted reference}
\label{app:base}
B1 fits a real channel map in each of 16 radial Fourier bands and then freezes it; B2, B2r and B3 use the same fitted reference:
\begin{equation}
\F(B_{\mathrm{spec}}(x))(k)=W_{b(k)}\F(z)(k).
\label{eq:app-base}
\end{equation}
Let $Z_i(k)=\F(z_i)(k)$ and $T_i(k)=\F(y_i)(k)$ for training pair $i$.
For $\Omega_b^+=\{k\in\Omega_+:b(k)=b\}$, define
\begin{align}
A_b&=\sum_{i,k\in\Omega_b^+}\operatorname{Re}\{Z_i(k)Z_i(k)^*\},\\
M_b&=\sum_{i,k\in\Omega_b^+}\operatorname{Re}\{T_i(k)Z_i(k)^*\},\\
\lambda_b&=10^{-6}\left(\frac{\tr A_b}{d}+10^{-12}\right),
\qquad d=C_{\mathrm{in}}+1 .
\label{eq:app-ridge}
\end{align}
Here $*$ is conjugate transpose and $W_b$ is real.

\begin{proposition}[Implemented fitting objective]
\label{prop:ridge}
For fixed training data, $W_b=M_b(A_b+\lambda_b I)^{-1}$ uniquely minimizes
\begin{equation}
Q_b(W)=\sum_{i,k\in\Omega_b^+}\lVert T_i(k)-WZ_i(k)\rVert_2^2
+\lambda_b\lVert W\rVert_F^2 .
\label{eq:ridge-objective}
\end{equation}
\end{proposition}
\paragraph{Proof.}
Expanding gives
$Q_b(W)=c-2\tr(WM_b^\top)+\tr(WA_bW^\top)+\lambda_b\tr(WW^\top)$,
where $c$ is independent of $W$.
The normal equation is $W(A_b+\lambda_b I)=M_b$.
Since $A_b$ is positive semidefinite and $\lambda_b>0$, the quadratic is strictly convex and the solution is unique.
\hfill$\square$

The implementation solves this linear system rather than explicitly computing its inverse.
The constant makes the fitted reference affine in $\mathcal U_s(x)$, although it is linear in $z$.
Every band excluding zero frequency has an identically zero final row and column in $A_b$, so the unregularized matrix is singular there.
Ridge resolves this singularity and selects zero for the unused offset coefficients, which matches the zero offset of Eq.~\eqref{eq:b0-fourier}.

\paragraph{Limits of the fitting guarantee.}
Proposition~\ref{prop:ridge} guarantees $Q_b(W_b)\le Q_b(W')$ for any competing matrix $W'$, including $[A,\mathbf 0]$.
It does not guarantee smaller unpenalized reconstruction error than URT's reference.
The residual cross-moment satisfies $M_b-W_bA_b=\lambda_bW_b$ rather than exact sample orthogonality.

\paragraph{Stored-bin weighting.}
For the default DFT normalization, Parseval requires
\begin{equation}
\lVert e\rVert_2^2=\frac{1}{HW}
\sum_{k\in\Omega_+}m(k)|\F(e)(k)|^2 ,
\label{eq:parseval}
\end{equation}
where $m(k)=1$ on the zero horizontal-frequency column and the Nyquist column when present, and $m(k)=2$ elsewhere.
Our implementation uses unit stored-bin weights and sums raw spectra without per-image size normalization.
Its fitting objective therefore differs from ordinary equal-image-weight pixel MSE.
These conventions are part of the estimator, not a claim of statistical optimality.

\subsection{Residual gains and calibration}
\label{app:gradients}
B2 initializes per-channel, per-band residual gains to one without a gain penalty.
B2r uses a random initial profile in $[0.05,1]$ as its penalty center.
B3 obtains its profile from two-fold cross-fitting on training data.
For complementary training groups $S_1,S_2$, fit $B^{(j)}_{\mathrm{spec}}$ on $S_j$ and define the held-out residual
$e_i^{(j)}=y_i-B^{(j)}_{\mathrm{spec}}(x_i)$ for $i\in S_{3-j}$.
For channel $c$ and band $b$, accumulate
\begin{align}
R_{c,b}&=\sum_{j=1}^2\sum_{i\in S_{3-j}}\sum_{k\in\Omega_b^+}
|\F(e_i^{(j)})_c(k)|^2,\\
T_{c,b}&=\sum_{j=1}^2\sum_{i\in S_{3-j}}\sum_{k\in\Omega_b^+}
|\F(y_i)_c(k)|^2.
\end{align}
With $\epsilon=10^{-12}$ and $\gamma=0.05$,
\begin{align}
\rho_{c,b}&=\clip\!\left(\frac{R_{c,b}}{\max(T_{c,b},\epsilon)},0,1\right),\\
g_{0,c,b}&=\max\!\left\{\gamma,
\frac{\sqrt{\rho_{c,b}}}{\max(\max_{b'}\sqrt{\rho_{c,b'}},\epsilon)}\right\}.
\label{eq:app-g0}
\end{align}
The normalization is per output channel, and the floor applies only at initialization.
After cross-fitting, the reference is refitted on all training pairs and frozen.
The gate acts on the target-grid residual $r=\mathcal P_s(f_\theta(x))$:
\begin{equation}
\F(\G_g r)_c(k)=g_{c,b(k)}\F(r)_c(k).
\label{eq:supp-app-gate}
\end{equation}
B2r and B3 minimize
\begin{equation}
\begin{aligned}
L_{\mathrm{data}}&=\frac1n\sum_i\meanabs{B_{\mathrm{spec}}(x_i)+\G_g r_i-y_i},\\
L(\theta,g)&=L_{\mathrm{data}}+\frac{\mu}{C_{\mathrm{out}}B}\lVert g-g_0\rVert_F^2,
\end{aligned}
\label{eq:app-gain-loss}
\end{equation}
with $B=16$ bands and $\mu=10^{-3}$.
B1 has no gate, and B2 has no penalty.
Gains are unconstrained during training.
B3 versus B2r compares measured and random penalty centers at the same strength; it does not isolate initialization from regularization.

Let $P_{c,b}r$ retain channel $c$ and band $b$ of a residual image.
The gate is $\G_g r=\sum_{c,b}g_{c,b}P_{c,b}r$.
Real radial gains preserve conjugate symmetry and define a self-adjoint operator on real images.

\begin{lemma}[Gate derivatives]
\label{lem:gradient}
At a differentiable point of the data loss, let $q=\partial L/\partial\hat y$.
Then
\begin{equation}
\frac{\partial L}{\partial r}=\G_gq,\qquad
\frac{\partial L}{\partial g_{c,b}}=\langle q,P_{c,b}r\rangle .
\label{eq:app-gate-gradient}
\end{equation}
The absolute loss uses a selected subgradient at nondifferentiable points.
\end{lemma}
\paragraph{Proof.}
For fixed gains, the derivative with respect to $r$ is $\G_g$, whose adjoint equals itself.
The derivative with respect to $g_{c,b}$ is the corresponding band projection of $r$.
The chain rule gives both identities.
\hfill$\square$

An exactly zero body output makes every data-loss gain gradient zero.
All five configurations initialize the body output weights with standard deviation $10^{-3}$ and zero bias.
For the gated variants, this permits nonzero initial gain gradients and starts the prediction near the fitted reference.
It does not guarantee a nonzero gradient in every band: an empty band, a zero projected residual or cancellation in the inner product can still give zero.
The penalty contributes $2\mu(g-g_0)/(C_{\mathrm{out}}B)$, which vanishes at $g=g_0$.

\paragraph{Exact identities and a local approximation.}
Lemmas~\ref{lem:commute} and~\ref{lem:gradient} and Proposition~\ref{prop:ridge} are exact.
The following is only a heuristic.
Let $J=\partial r/\partial\theta$.
Holding gains fixed, a locally linearized plain-gradient body step with rate $\eta$ gives
\begin{equation}
\Delta\hat y\approx-\eta\,\G_gJJ^\top\G_gq ,
\label{eq:local-step}
\end{equation}
and only under the further approximation $JJ^\top\approx\kappa I$ does this become $-\eta\kappa\G_g^2q$.
The implementation uses Adam and updates the body and gate jointly, so Eq.~\eqref{eq:local-step} describes a mechanism, not the training update.

\paragraph{Meaning of the calibration statistic.}
Out-of-fold error measures the prediction error of a restricted fitted reference.
It includes approximation error, finite-sample error, unpredictable content, preprocessing effects and noise, and does not identify irreducible sensor uncertainty.
Clipping the ratio at one discards magnitude differences among larger errors.
Grouped cross-fitting avoids scoring a fitted reference on the fold used to fit it, but does not prove unbiased risk estimation for the final full-data model.

\paragraph{Tile boundaries.}
RealSR inference uses overlapping low-resolution tiles (Section~\ref{app:protocols}).
For URT the tiling only affects the body, because $B_s$ is computed pointwise after bicubic interpolation.
For the spectral variants, the Fourier operations of the fitted reference and the gate act within each tile, so full-image calibration, crop training and tilewise inference impose different boundaries.

\subsection{NAFNet body and output initialization}
\label{app:body}
The body $f_\theta$ follows the NAFNet block of Chen et al.~\cite{nafnet}.
Let $\mathrm{LN}$ be channel-wise layer normalization, $\mathrm{DW}$ a $3\times3$ depthwise convolution, and $\mathrm{SG}(u)=u_{1:c}\odot u_{c+1:2c}$ the SimpleGate that halves $2c$ channels.
With $\mathrm{SCA}(v)=\mathrm{Conv}_{1\times1}(\mathrm{GAP}(v))$ (simplified channel attention; GAP is global average pooling), one block maps a $c$-channel feature $u$ to
\begin{align}
v&=\mathrm{SG}\!\left(\mathrm{DW}\!\left(\mathrm{Conv}_{1\times1}^{c\to2c}(\mathrm{LN}(u))\right)\right),\\
u'&=u+\alpha_1\odot\mathrm{Conv}_{1\times1}\!\left(v\odot\mathrm{SCA}(v)\right),\\
\mathrm{blk}(u)&=u'+\alpha_2\odot\mathrm{Conv}_{1\times1}^{c\to c}\!\left(\mathrm{SG}\!\left(\mathrm{Conv}_{1\times1}^{c\to2c}(\mathrm{LN}(u'))\right)\right),
\label{eq:naf-block}
\end{align}
where the per-channel scales $\alpha_1,\alpha_2\in\mathbb R^{c}$ are initialized to zero, so every block is the identity at initialization.
After bottom and right reflection padding to a multiple of 16, a $3\times3$ convolution lifts $x$ to $w$ channels.
Encoder level $\ell\in\{1,\dots,4\}$ applies $(1,1,1,28)$ blocks at width $2^{\ell-1}w$, stores the result as a skip feature, and halves the resolution with a stride-2 $2\times2$ convolution that doubles the width.
One block acts at width $16w$.
Each decoder level applies a $1\times1$ convolution that doubles the channels, a pixel shuffle $\times2$, addition of the skip feature, and one block.
A $3\times3$ output convolution with bias maps the final $w$-channel feature $h$ to $C_{\mathrm{out}}s^2$ channels, and the result is cropped to $H\times W$.

\begin{proposition}[Initial correction]
\label{prop:init}
Condition on $x$ and on every parameter other than the output-convolution weights $W$, so that $h$ is fixed.
Let the entries of $W$ be independent $\mathcal N(0,\sigma^2)$ and the bias zero.
Then the output at channel $c$ and position $(i,j)$ is $o_c(i,j)=\sum_{k,u,v}W_{c,k,u,v}\,h_k(i+u,j+v)$ (zero padding outside the feature map), and
\begin{equation}
\E[o_c(i,j)\mid h]=0,\qquad
\E[o_c(i,j)^2\mid h]=\sigma^2\sum_{k,u,v}h_k(i+u,j+v)^2 .
\label{eq:supp-init-moments}
\end{equation}
Outputs in different channels are uncorrelated given $h$.
\end{proposition}
\paragraph{Proof.}
The output is linear in $W$ with fixed coefficients $h_k(i+u,j+v)$.
Linearity and $\E W=0$ give the mean, and independence of the weights gives the variance.
Weights of different output channels are independent with zero mean, so the cross-moment vanishes.
\hfill$\square$

The proposition is exact but conditional on $h$.
It does not bound the initial residual deterministically and makes no claim about convergence speed or final accuracy.

\section{Experimental details}
\label{app:experimental}
This section records the settings needed to reproduce the reported numbers.
Section~\ref{app:datasets} gives the dataset splits and preprocessing, Section~\ref{app:protocols} the URT and spectral-variant training settings, and Section~\ref{app:retrained} the baseline runs.
Section~\ref{app:scorers} defines the scorers, Section~\ref{app:hardware} the hardware and software, Section~\ref{app:efficiency} the latency and parameter counts, and Section~\ref{app:selection} how checkpoints and configurations were selected.

\subsection{Datasets}
\label{app:datasets}
The splits and preprocessing are those of main-paper Section~\ref{sec:protocol}; this subsection adds the preprocessing details not given there.
\paragraph{IXI preprocessing.}
T2 is affinely registered onto T1 using mutual information.
Each volume is divided by its own mean, and scaling maxima are fitted on training subjects only.
The input window selects brain slices below the vertex and stacks five neighboring T1 slices.
Evaluation separately mean-normalizes prediction and reference, then divides both by the reference maximum.
This normalization uses the reference image only at scoring time; no preprocessing statistic is fitted on test subjects.

\paragraph{LLVIP.}
Training and validation come from the official training directory, and the test set is the official test directory (scenes 19 to 26, 3,463 pairs).
Validation uses every fifth frame of training scenes 07, 11 and 15 (305 pairs); all frames of these three scenes are excluded from training, which leaves 10,502 training pairs.
Holding out whole scenes avoids selecting checkpoints on near-duplicates of training frames.
Frames are resized from $1280\times1024$ to $256\times256$ with bicubic interpolation, without preserving the aspect ratio.
The infrared target is stored as an eight-bit three-channel image; training uses its channel mean as the one-channel target $y$.

\paragraph{SEN1-2.}
A scene is one season and region-of-interest folder.
The 209/23/26 training/validation/test scenes contribute at most 40 patches each, evenly spaced over the scene's patch numbers, which gives 8,360/920/1,040 pairs at the native $256\times256$ resolution.
SAR is decoded as one eight-bit channel and the optical target as eight-bit RGB.

\paragraph{RealSR.}
Pairs are the Canon and Nikon images of the official RealSR V3 training and test folders at each scale, matched by file name.
The high-resolution image is cropped to $s$ times the low-resolution size; pairs whose sizes differ by more than $s$ pixels are skipped.
This gives 406, 459 and 400 training pairs at $\times2$, $\times3$ and $\times4$, and 100 test pairs (50 Canon, 50 Nikon) at each scale.
Training draws random $48\times48$ low-resolution crops with the matching high-resolution crop and one of the eight flips and right-angle rotations.

\subsection{URT and spectral-variant configurations}
\label{app:protocols}
Table~\ref{tab:settings} records the common task settings.
For the spectral variants, the two-fold partition alternates sorted training subject or scene identifiers; RealSR alternates image indices.

\begin{table*}[t]
\centering\small
\caption{URT and spectral-variant training settings.
Widths denote feature channels, not matched parameter budgets.
LR denotes low resolution.}
\label{tab:settings}
\begin{tabular}{@{}lrrrrl@{}}
\toprule
Task & Input/output channels & Scale & Width & Batch & Initial learning rate\\
\midrule
IXI & $5/1$ & 1 & 48 & 8 & $10^{-4}$\\
LLVIP & $3/1$ & 1 & 32 & 8 & $2\cdot10^{-4}$\\
SEN1-2 & $1/3$ & 1 & 32 & 8 & $2\cdot10^{-4}$\\
RealSR $\times2$ & $3/3$ & 2 & 32 & 32 & $10^{-4}$\\
\bottomrule
\end{tabular}
\end{table*}

All final residual-model runs use seed $0$ and L1 supervision.
Adam uses $(0.9,0.999)$ moments and cosine decay to $10^{-7}$.
IXI uses flips and affine augmentation with probability $0.7$; LLVIP uses flips; SEN1-2 uses flips and right-angle rotations; RealSR uses $48\times48$ LR crops with flips and rotations.
The RealSR $\times3$ and $\times4$ runs use the same body, objective and $48\times48$ LR crops with scale-specific output heads.
Translation inference uses flip averaging.
RealSR uses overlapping LR tiles of size $256$ and overlap $24$, without flip or rotation averaging.

\subsection{Baselines}
\label{app:retrained}
Table~\ref{tab:baselines} lists the 22 baseline runs of the main comparison.
Every baseline is scored with the same scorer as URT.

\begin{table*}[t]
\centering\small
\caption{Baseline runs in the main comparison. Implementation: official code; BasicSR architecture trained with URT's $\ell_1$ loss and Adam schedule; reimplementation from the paper ($\dagger$ in the main table); or a local implementation. Checkpoint: the validation-PSNR maximum for translation and the last iterate for RealSR, as for URT.}
\label{tab:baselines}
\setlength{\tabcolsep}{4pt}
\begin{tabularx}{\textwidth}{@{}llll>{\raggedright\arraybackslash}X@{}}
\toprule
Setting & Model & Implementation & Supervision & Input and deviations\\
\midrule
IXI & pix2pix & Official & Paired & Five-slice input; default loss and optimizer\\
IXI & SynDiff & Official & Unpaired objective & Single T1 slice, the central slice of URT's five-slice stack\\
LLVIP & pix2pix & Official & Paired & Default loss and optimizer\\
LLVIP & CycleGAN & Official & Unpaired & Default loss and optimizer\\
LLVIP & BCI (Pyramid pix2pix) & Official & Paired & Gaussian-pyramid supervision as published\\
LLVIP & Consistency model & Official & Paired & One-step sampling\\
LLVIP & CMF & Local & Paired & One-step sampling with a fixed noise seed\\
SEN1-2 & S2O-TDN & Reimplementation & Paired & Pixel-level objective; VGG perceptual term not used\\
SEN1-2 & CMF & Local & Paired & One-step sampling with a fixed noise seed\\
RealSR $\times2$ & EDSR, RCAN & BasicSR & Paired & EDSR at baseline size\\
RealSR $\times2$ & OR-Net & Reimplementation & Paired & None\\
RealSR $\times2$ & MPF-Net & Reimplementation & Paired & $\ell_1$ objective\\
RealSR $\times2$ & CMF & Local & Paired & One-step sampling with a fixed noise seed\\
RealSR $\times3,\times4$ & EDSR, RCAN & BasicSR & Paired & EDSR at baseline size\\
RealSR $\times3,\times4$ & OR-Net & Reimplementation & Paired & None\\
RealSR $\times3,\times4$ & MPF-Net & Reimplementation & Paired & $\ell_1$, then $\ell_1$ plus contrastive regularization with an EDSR teacher at the same scale\\
\bottomrule
\end{tabularx}
\end{table*}

pix2pix and CycleGAN keep their default losses and optimizer, and SynDiff keeps its official unpaired objective.
MPF-Net uses different objectives at $\times2$ and at $\times3$ and $\times4$ (Table~\ref{tab:baselines}).

\subsection{Scorers}
\label{app:scorers}
All scores are computed per test image (per slice for IXI) and averaged.
Predictions are clipped to $[0,1]$ before scoring.
\paragraph{IXI.} PSNR and SSIM follow the normalization of Section~\ref{app:datasets}; SSIM uses the default $7\times7$ uniform window with data range one, reported as a fraction.
LPIPS (AlexNet) uses the flip-averaged, clipped prediction and the target in $[0,1]$, repeated to three channels and mapped to $[-1,1]$, over the 1,000 test slices.
Slice dispersion is not a confidence interval over subjects or seeds.
\paragraph{LLVIP.} PSNR and SSIM use the floating-point prediction replicated to three channels and the stored three-channel target, with SSIM averaged over channels.
LPIPS and FID use the prediction quantized to eight bits and replicated to three channels against the stored eight-bit target.
\paragraph{SEN1-2.} Predictions are quantized to eight bits before every metric; PSNR and SSIM are computed over RGB.
\paragraph{RealSR.} Predictions are quantized to eight bits.
PSNR-Y and SSIM-Y use the BT.601 luminance of MATLAB's \texttt{rgb2ycbcr} and remove a border of $s$ pixels; SSIM-Y uses a Gaussian window with standard deviation $1.5$ and population covariance.
PSNR-RGB uses all three channels without border removal, and LPIPS uses the full RGB images.
\paragraph{Common components.} LPIPS uses AlexNet features with inputs mapped to $[-1,1]$.
FID uses the $2048$-dimensional Inception pool features of the TorchMetrics implementation on eight-bit images.
These conventions differ from those of some published tables and should be taken into account in cross-paper comparisons.

\subsection{Hardware and software}
\label{app:hardware}
Every URT and spectral-variant run, its evaluation and the efficiency profile used one NVIDIA Tesla P100 with 16\,GB, in fp32.
The retrained baselines ran on one or two NVIDIA Tesla P100 (16\,GB) or A100 (80\,GB) GPUs.
The software is Python 3.11 and PyTorch 2.5.1 with CUDA 12.1.
Other libraries are torchvision 0.20.1, NumPy 2.2.6, SciPy 1.14.1, scikit-image 0.25.2, SimpleITK 2.4.0, Pillow 12.2.0, lpips 0.1.4 and TorchMetrics 1.4.3.

\subsection{Efficiency}
\label{app:efficiency}
We profiled latency for B3, which runs the URT residual body and adds the fitted reference and gain operations, so its latency includes every operation of URT.
The profile measures a single forward pass without flip averaging or tiling: batch size one, fp32, a $256\times256$ input (the low-resolution input for RealSR $\times2$), 10 warm-up passes, then the mean of 50 timed passes.
The latencies are 41.39\,ms for IXI, 41.37\,ms for LLVIP, 41.18\,ms for SEN1-2 and 40.66\,ms for RealSR $\times2$.
Full translation inference evaluates the body twice for flip averaging (main Eq.~\eqref{eq:b0-infer}), and RealSR inference runs one pass per tile.
The URT parameter counts of the main comparison are 38.292\,M for IXI, 17.111\,M for LLVIP and SEN1-2, and 17.115, 17.119 and 17.125\,M for RealSR $\times2$, $\times3$ and $\times4$; they differ only through the input stem, the output head and the width.

\subsection{Selection criteria and outcomes}
\label{app:selection}
Training and calibration use training examples.
Translation checkpoints follow validation PSNR; RealSR uses a fixed final endpoint without a validation set.
During development we compared several candidate runs and inference modes before fixing the final configurations.
Within the residual family, the spectral variants B1 to B3 were evaluated as alternatives to URT (Section~\ref{app:family}).
We report URT as the main configuration because it has the best mixed mean rank in main Table~\ref{tab:variants}(c), which combines translation validation PSNR with RealSR $\times2$ test PSNR-Y.

\section{Additional results}
\label{app:results}
This section reports results that complement the main comparison.
Sections~\ref{app:family} and~\ref{app:sensitivity} break the residual-family study down by task and by single changed setting.
Sections~\ref{app:rgb} and~\ref{app:paired} report RealSR in RGB and a per-camera paired comparison with RCAN, and Section~\ref{app:worked} examines an image on which PSNR and LPIPS disagree.
Sections~\ref{app:urt-qual} and~\ref{app:variant-plots} give qualitative examples and summary plots.
Every model is a single training run, scored with the protocol of Section~\ref{app:experimental}.

\subsection{Residual-family changes by task}
\label{app:family}
Table~\ref{tab:family-delta} expresses main Table~\ref{tab:variants}(a) as test PSNR changes relative to URT.
No configuration improves on URT in every task: the fitted reference alone (B1) helps only on SEN1-2, every gated variant helps on LLVIP, and all four variants are lower on IXI and RealSR $\times2$.
The cross-fitted profile (B3) is below the unit profile (B2) on IXI, LLVIP and RealSR $\times2$.

\begin{table}[t]
\centering\small
\caption{Test PSNR change (dB) of each spectral variant relative to URT (B0). RealSR uses PSNR-Y at $\times2$. Single runs; differences are not adjusted for run-to-run variation.}
\label{tab:family-delta}
\begin{tabular}{@{}lrrrr@{}}
\toprule
Variant & IXI & LLVIP & SEN1-2 & RealSR $\times2$\\
\midrule
B1 & $-0.096$ & $-0.167$ & $+0.189$ & $-0.129$\\
B2 & $-0.111$ & $+0.229$ & $+0.013$ & $-0.107$\\
B2r & $-0.144$ & $+0.206$ & $-0.072$ & $-4.145$\\
B3 & $-0.385$ & $+0.191$ & $+0.042$ & $-0.336$\\
\midrule
URT (B0) & 30.774 & 13.126 & 13.478 & 33.596\\
\bottomrule
\end{tabular}
\end{table}

\subsection{Single-setting contrasts}
\label{app:sensitivity}
Three pairs of residual-family runs differ in exactly one setting and share the data split, body, width, schedule, seed, checkpoint rule and scorer (Table~\ref{tab:contrasts}).
B1 against URT changes only the reference, from the fixed map $B_s$ to the fitted spectral map.
B2 against B1 adds unit-initialized bandwise gains without a gain penalty.
B3 against B2r replaces the random initial gain profile and penalty center by the measured cross-fitted profile at the same penalty strength.
The comparison of B2 with B2r, and of B3 with B2, changes both the initial profile and the penalty, so it is not a single-setting contrast.
The sign of a contrast depends on the task: the fitted reference helps only on SEN1-2, the unit gains help most on LLVIP, and the measured profile has a large positive contrast on RealSR $\times2$ only because the random profile (B2r) is far below the other configurations there (Table~\ref{tab:family-delta}).
These contrasts vary the reference and the gains while keeping the learned body; they compare residual designs rather than isolating the body.

\begin{table*}[t]
\centering\small
\caption{Single-setting contrasts among the residual-family runs: change in test PSNR (dB) and test SSIM from the first to the second run of each pair. RealSR is PSNR-Y and SSIM-Y at $\times2$; IXI SSIM is shown as a fraction. Single runs with seed $0$; the differences are not adjusted for run-to-run variation.}
\label{tab:contrasts}
\setlength{\tabcolsep}{4pt}
\begin{tabular}{@{}>{\raggedright\arraybackslash}p{.34\textwidth}rrrrrrrr@{}}
\toprule
 & \multicolumn{4}{c}{PSNR change (dB)} & \multicolumn{4}{c}{SSIM change}\\
\cmidrule(lr){2-5}\cmidrule(l){6-9}
Contrast (setting changed) & IXI & LLVIP & SEN1-2 & RealSR $\times2$ & IXI & LLVIP & SEN1-2 & RealSR $\times2$\\
\midrule
B1 $-$ URT (fixed reference replaced by fitted reference) & $-0.096$ & $-0.167$ & $+0.189$ & $-0.129$ & $-0.017$ & $-0.006$ & $-0.001$ & $-0.006$\\
B2 $-$ B1 (unit gains added) & $-0.014$ & $+0.397$ & $-0.176$ & $+0.022$ & $-0.001$ & $+0.013$ & $+0.007$ & $+0.001$\\
B3 $-$ B2r (measured instead of random gain profile) & $-0.240$ & $-0.015$ & $+0.114$ & $+3.809$ & $-0.001$ & $+0.005$ & $+0.001$ & $+0.151$\\
\bottomrule
\end{tabular}
\end{table*}

\subsection{RealSR in RGB}
\label{app:rgb}
Table~\ref{tab:crst-rgb} reports full-color PSNR at the same fixed endpoints as the main luminance results.
URT has the highest RGB PSNR at $\times3$ and $\times4$, 0.224 and 0.229\,dB above the next model (OR-Net), and the highest among the five residual configurations at $\times2$, where OR-Net, EDSR and MPF-Net are higher by at most 0.397\,dB.

\begin{table}[t]
\centering\small
\caption{RealSR test PSNR-RGB (dB, higher is better) at the endpoints of the main comparison. No validation selection is performed. ``n/a'' marks settings without a run.}
\label{tab:crst-rgb}
\begin{tabular}{@{}lrrr@{}}
\toprule
Model & $\times2$ & $\times3$ & $\times4$\\
\midrule
URT (ours) & 31.763 & \textbf{27.085} & \textbf{25.725}\\
EDSR & \underline{32.009} & 26.263 & 24.632\\
RCAN & 31.606 & 26.540 & 25.132\\
OR-Net & \textbf{32.160} & \underline{26.861} & \underline{25.496}\\
MPF-Net & 31.920 & 23.862 & 22.330\\
CMF & 29.033 & n/a & n/a\\
\midrule
\quad B1 & 31.607 & n/a & n/a\\
\quad B2 & 31.626 & n/a & n/a\\
\quad B2r & 27.823 & n/a & n/a\\
\quad B3 & 31.379 & n/a & n/a\\
\bottomrule
\end{tabular}
\end{table}

\subsection{Paired RealSR comparison by camera}
\label{app:paired}
URT and RCAN were scored on the same 100 test images at each of $\times3$ and $\times4$, matched by image identifier.
Table~\ref{tab:paired} summarizes the per-image PSNR-Y differences.
The mean gap equals the difference of the main-table means. URT is higher on average at both scales and on both cameras except Canon at $\times3$ ($-0.018$\,dB, well within one standard deviation of $0.268$\,dB); the gain grows with scale, from $+0.058$ to $+0.164$\,dB, and URT is higher on 74 of the 100 images at $\times4$.
The spread describes variation across test images for one trained model per method, not variation across training seeds.

\begin{table}[t]
\centering\small
\caption{Per-image PSNR-Y difference URT minus RCAN (dB) on the RealSR test set. SD is the standard deviation across images; ``URT higher'' counts images with a positive difference.}
\label{tab:paired}
\setlength{\tabcolsep}{4pt}
\begin{tabular}{@{}llrrrr@{}}
\toprule
Scale & Camera & $n$ & Mean & SD & URT higher\\
\midrule
$\times3$ & Canon & 50 & $-0.018$ & 0.268 & 24\\
 & Nikon & 50 & $+0.134$ & 0.308 & 32\\
 & All & 100 & $+0.058$ & 0.297 & 56\\
\midrule
$\times4$ & Canon & 50 & $+0.070$ & 0.237 & 33\\
 & Nikon & 50 & $+0.259$ & 0.360 & 41\\
 & All & 100 & $+0.164$ & 0.318 & 74\\
\bottomrule
\end{tabular}
\end{table}

\subsection{When PSNR and LPIPS disagree}
\label{app:worked}
Per-image PSNR-Y and LPIPS often rank URT and RCAN differently: they disagree on 32, 44 and 53 of the 100 test images at $\times2$, $\times3$ and $\times4$.
To show one such case, we scanned the displayed RealSR items (Nikon\_002, Nikon\_014, Nikon\_034) in their fixed order, from $\times2$ to $\times4$, and took the first item on which the two metrics rank URT and RCAN differently.
This is Nikon\_002 at $\times2$ (first row of Figure~\ref{fig:urt-sr}): URT has lower PSNR-Y than RCAN (32.262 versus 32.309\,dB) but lower LPIPS (0.0962 versus 0.1013).
The same scene at $\times4$ shows the opposite pattern (first row of Figure~\ref{fig:urt-sr-x4}): URT has higher PSNR-Y (26.175 versus 25.902\,dB) and higher LPIPS (0.4097 versus 0.3851).
All four scores are computed on the full test image, not on the displayed crop; the crop is the target-guided window used in every column.
At both scales the two metrics order URT and RCAN differently, in opposite directions, so a PSNR-Y advantage does not imply an LPIPS advantage; the main paper therefore reports both metrics.

\subsection{Qualitative examples}
\label{app:urt-qual}
\begin{figure*}[p]
\centering
\includegraphics[width=\textwidth,height=.8\textheight,keepaspectratio]{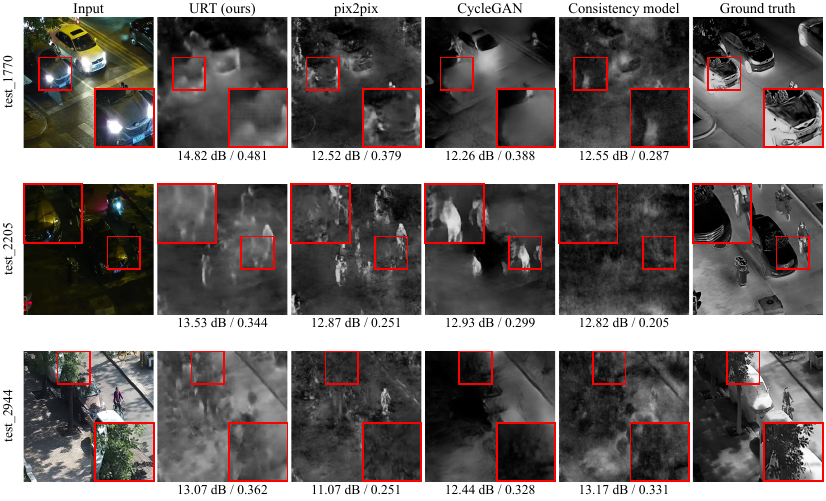}
\caption{URT (ours) and baseline models on LLVIP test frames 1770, 2205 and 2944. Columns: visible input, URT (ours), pix2pix, CycleGAN, consistency model, and the infrared ground truth.
All output columns use the fixed $[0,1]$ intensity range without per-image stretching. Values are per-example PSNR (dB) and fractional SSIM on floating-point predictions, as in the main LLVIP table.}
\label{fig:urt-llvip}
\end{figure*}

\begin{figure*}[p]
\centering
\includegraphics[width=\textwidth,height=.8\textheight,keepaspectratio]{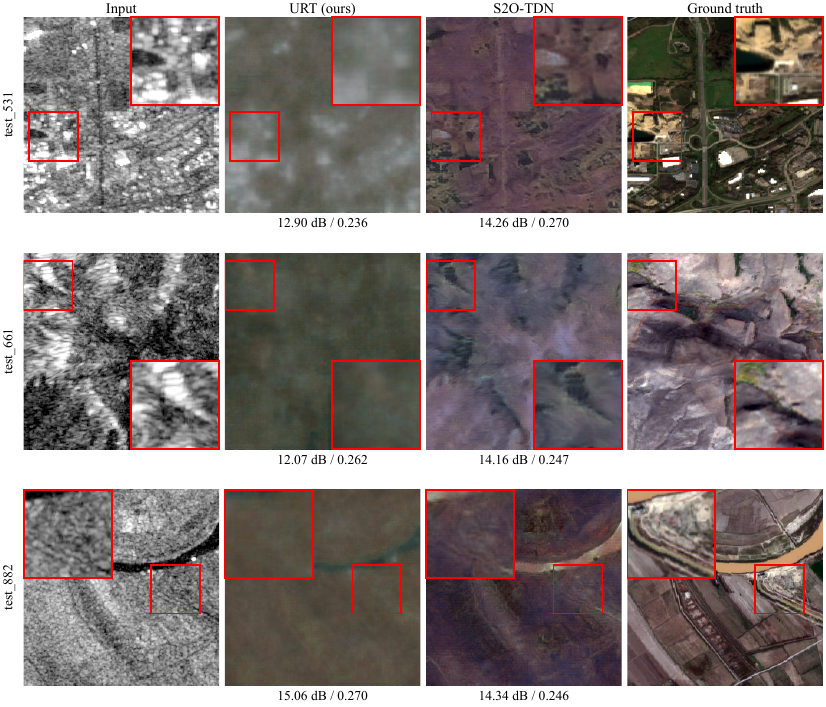}
\caption{URT (ours) and the S2O-TDN baseline on SEN1-2 test patches 531, 661 and 882. Columns: SAR input (also the fixed SAR-to-RGB reference), URT (ours), S2O-TDN, and the optical ground truth.
Predictions are quantized to eight bits as for scoring and shown in the fixed $[0,1]$ range. Values are per-example PSNR (dB) and fractional SSIM, as in the main SEN1-2 table.}
\label{fig:urt-sen12}
\end{figure*}

\begin{figure*}[p]
\centering
\includegraphics[width=\textwidth,height=.8\textheight,keepaspectratio]{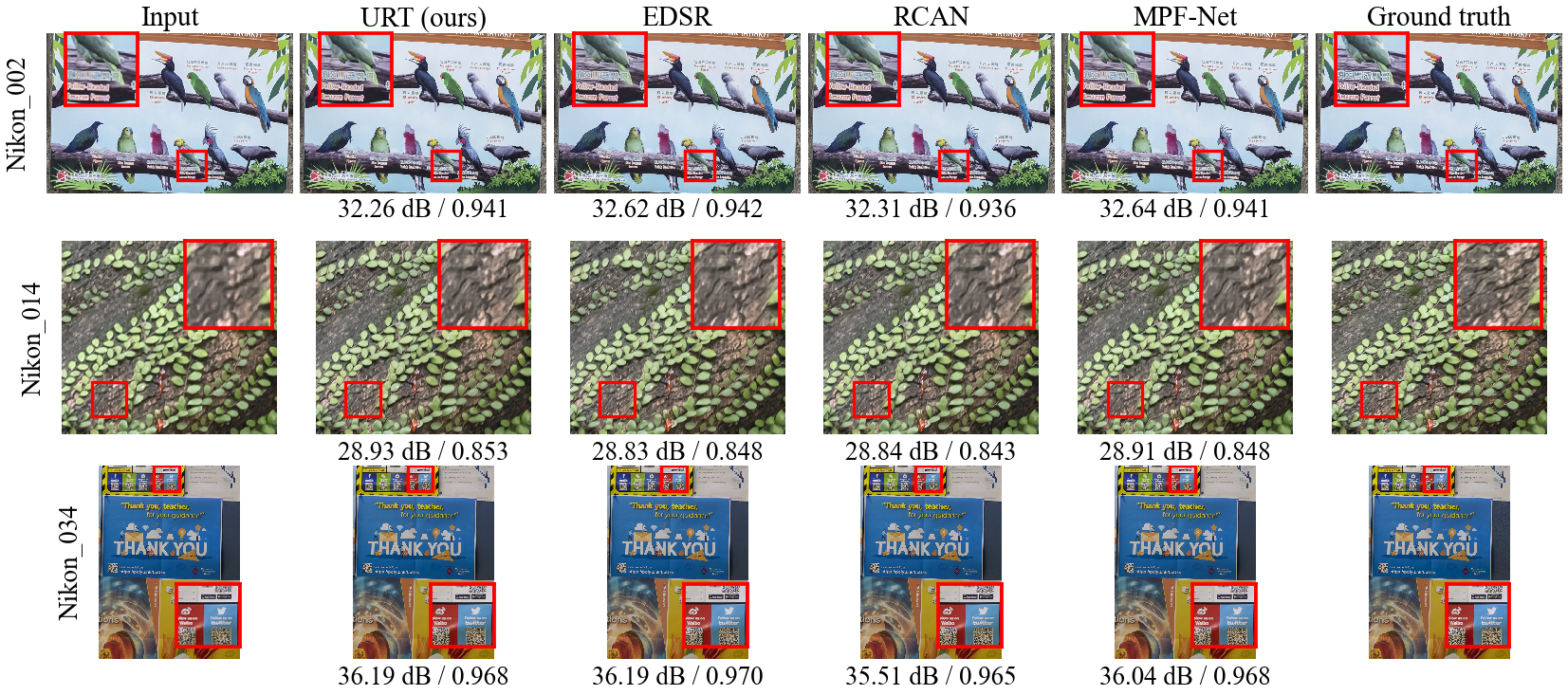}
\caption{URT (ours) and baseline models on RealSR $\times2$ test images Nikon\_002, Nikon\_014 and Nikon\_034. Columns: the low-resolution input replicated $2\times2$ (nearest neighbor) for display only, URT (ours), EDSR, RCAN, MPF-Net, and the high-resolution ground truth.
Values are per-example PSNR-Y (dB) and SSIM-Y after eight-bit quantization and removal of a 2-pixel border, as in the main RealSR table.}
\label{fig:urt-sr}
\end{figure*}

\begin{figure*}[p]
\centering
\includegraphics[width=\textwidth,height=.8\textheight,keepaspectratio]{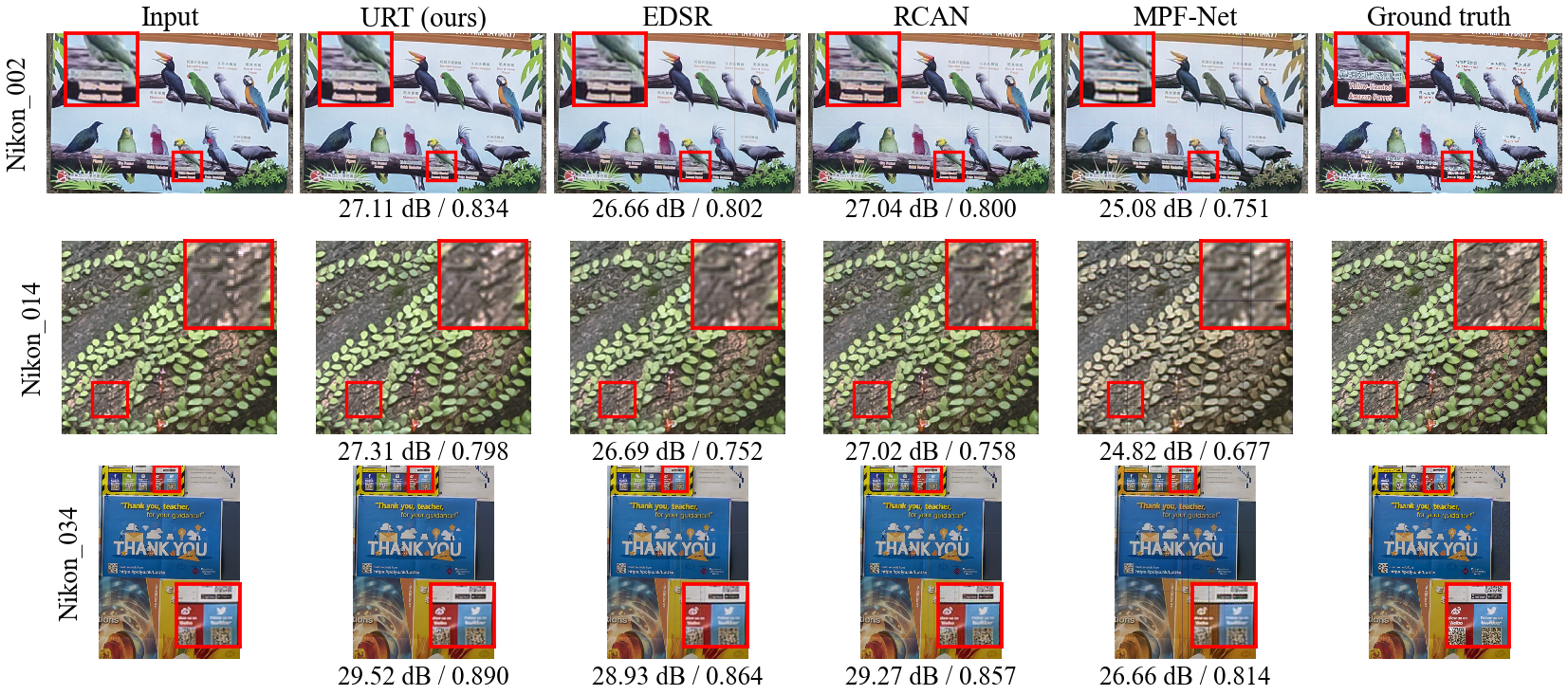}
\caption{URT (ours) and baseline models on RealSR $\times3$ test images Nikon\_002, Nikon\_014 and Nikon\_034. Columns: nearest-neighbor input, URT (ours), EDSR, RCAN, MPF-Net, and the high-resolution ground truth. The models are the $\times3$ endpoints of the main comparison. Values are per-example PSNR-Y (dB) and SSIM-Y after eight-bit quantization and removal of a 3-pixel border.}
\label{fig:urt-sr-x3}
\end{figure*}

\begin{figure*}[p]
\centering
\includegraphics[width=\textwidth,height=.8\textheight,keepaspectratio]{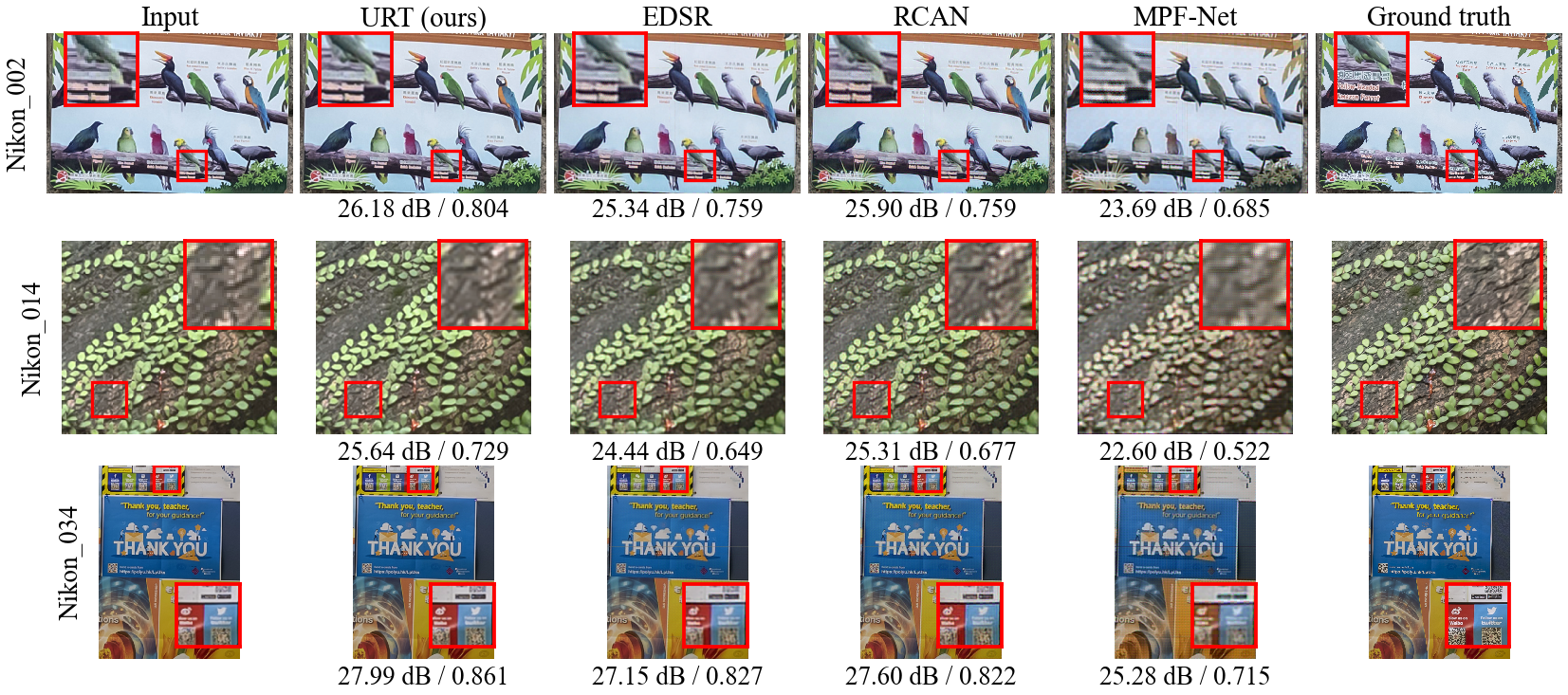}
\caption{URT (ours) and baseline models on RealSR $\times4$ test images Nikon\_002, Nikon\_014 and Nikon\_034. Columns: nearest-neighbor input, URT (ours), EDSR, RCAN, MPF-Net, and the high-resolution ground truth. The models are the $\times4$ endpoints of the main comparison. Values are per-example PSNR-Y (dB) and SSIM-Y after eight-bit quantization and removal of a 4-pixel border.}
\label{fig:urt-sr-x4}
\end{figure*}

Figures~\ref{fig:urt-llvip}, \ref{fig:urt-sen12}, \ref{fig:urt-sr}, \ref{fig:urt-sr-x3} and \ref{fig:urt-sr-x4} follow the main paper's IXI comparison layout for the other tasks and RealSR scales.
Examples are three randomly drawn test items, not selected by URT performance; for RealSR all three are Nikon images, and Table~\ref{tab:paired} reports both cameras.
Red boxes mark the most detailed window of the target (a target-guided rule), enlarged in the inset; the same window is used in every column.
No dataset-level number comes from these figures.

\subsection{Summary plots}
\label{app:variant-plots}
\begin{figure*}[t]
\centering
\includegraphics[width=\textwidth]{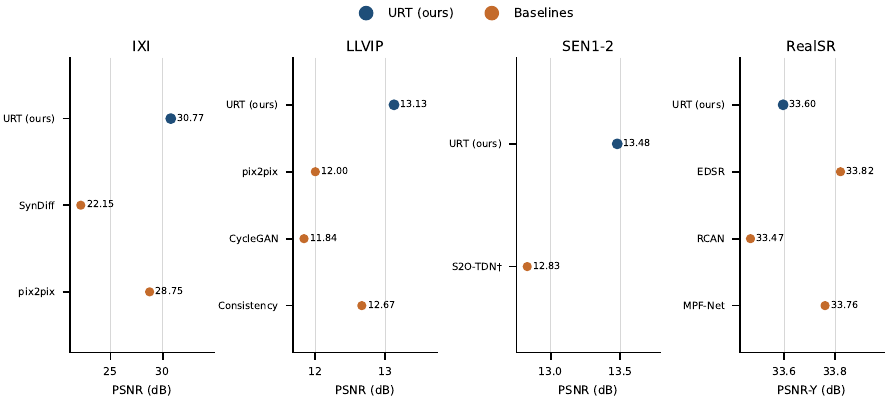}
\caption{Test PSNR (dB, higher is better) of URT and the baseline models of the main comparison. RealSR is PSNR-Y; baseline rows are shown only for the task on which they were evaluated.}
\label{fig:variants-psnr}
\end{figure*}

\begin{figure*}[t]
\centering
\includegraphics[width=\textwidth]{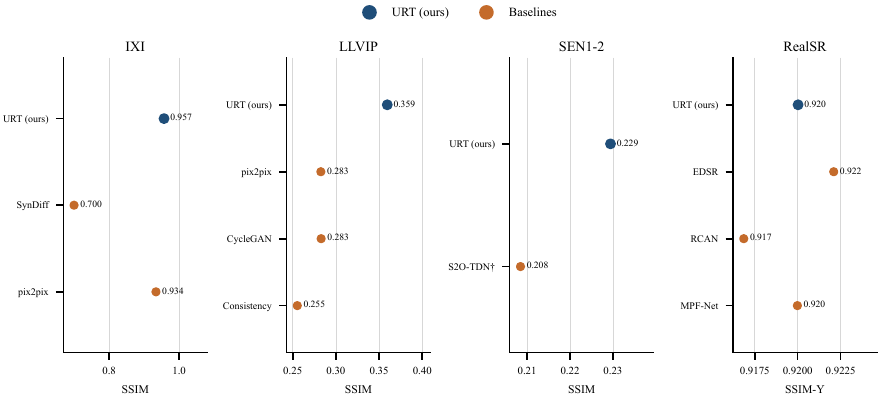}
\caption{Test SSIM of URT and the baseline models of the main comparison (fraction; RealSR is SSIM-Y). Baseline rows are shown only for the task on which they were evaluated.}
\label{fig:variants-ssim}
\end{figure*}

Figures~\ref{fig:variants-psnr} and~\ref{fig:variants-ssim} plot test PSNR and SSIM for URT and the baseline models of the main comparison on each task.
Each point is a single-seed run; no error bars are drawn.
Axes are zoomed to each task's range, so horizontal distances are not comparable across panels.

\section{Scope and limitations}
\label{app:scope}
\paragraph{Statistical evidence.}
The main comparison reports URT and 22 baseline runs evaluated with shared task data and scorers; the residual-family study reports five configurations on four settings.
Each dataset has separately trained weights, and every reported model is a single training run; Section~\ref{app:paired} reports per-image variation for the closest RealSR comparison.
URT and the spectral variants use seed $0$.


\paragraph{Scope of the claims.}
All tasks use paired, spatially aligned inputs; the results do not address unpaired or misaligned settings.
The identities in Section~\ref{app:derivations} specify the implemented models; they do not establish clinical utility or rule out hallucinated content.


\end{document}